\documentclass{ecai}
\usepackage{graphicx}
\usepackage{latexsym}
\usepackage{float}
\usepackage[small]{caption}
\usepackage{subcaption}
\usepackage[switch]{lineno}
\usepackage[figuresright]{rotating}

\usepackage{latexsym}
\usepackage{amssymb}
\usepackage{amsmath}
\usepackage{amsthm}
\usepackage{booktabs}
\usepackage{enumitem}
\usepackage{graphicx}
\usepackage{color}
\usepackage{algorithm}
\usepackage{algorithmic}
\usepackage{soul}
\usepackage{diagbox}
\usepackage{tabularx}
\usepackage{tcolorbox}

\begin{document}

\begin{frontmatter}



\title{Mind the Model, Not the Agent: The Primacy Bias in Model-based RL}


\author[A]{\fnms{Zhongjian}~\snm{Qiao}}
\author[A]{\fnms{Jiafei}~\snm{Lyu}}
\author[A]{\fnms{Xiu}~\snm{Li}\thanks{Corresponding Author. Email: li.xiu@sz.tsinghua.edu.cn.}} 

\address[A]{Tsinghua Shenzhen International Graduate School, Tsinghua University}


\begin{abstract}
The primacy bias in model-free reinforcement learning (MFRL), which refers to the agent's tendency to overfit early data and lose the ability to learn from new data, can significantly decrease the performance of MFRL algorithms. Previous studies have shown that employing simple techniques, such as resetting the agent's parameters, can substantially alleviate the primacy bias in MFRL. However, the primacy bias in model-based reinforcement learning (MBRL) remains unexplored. In this work, we focus on investigating the primacy bias in MBRL. We begin by observing that resetting the agent's parameters harms its performance in the context of MBRL. We further find that the primacy bias in MBRL is more closely related to the primacy bias of the world model instead of the primacy bias of the agent. Based on this finding, we propose \textit{world model resetting}, a simple yet effective technique to alleviate the primacy bias in MBRL. We apply our method to two different MBRL algorithms, MBPO and DreamerV2. We validate the effectiveness of our method on multiple continuous control tasks on MuJoCo and DeepMind Control Suite, as well as discrete control tasks on Atari 100k benchmark. The experimental results show that \textit{world model resetting} can significantly alleviate the primacy bias in the model-based setting and improve the algorithm's performance. We also give a guide on how to perform \textit{world model resetting} effectively.
\end{abstract}

\end{frontmatter}

\section{Introduction}
Deep Reinforcement Learning (DRL) has shown great potential in numerous fields such as robot control \cite{brunke2022safe} and autonomous driving \cite{kiran2021deep}. However, training an agent with good performance in DRL requires a large amount of data. Gathering samples from real environment can be costly, making it crucial to improve the algorithm's sample efficiency and train a better-performing agent using limited data. A direct approach to improving sample efficiency is to perform multiple training updates per environment step \cite{chen2021randomized,hiraoka2021dropout}, which means using a high UTD (Update-To-Data) ratio. This is analogous to updating multiple epochs for a dataset in supervised learning. However, unlike supervised learning, in DRL, the agent continuously interacts with the environment and learns from the replay buffer, which contains a dynamic and evolving data collection. As the policy is updated, the data distribution in the replay buffer also changes. In that case, using a high UTD ratio for updates can magnify the primacy bias \cite{nikishin2022primacy}, which refers to the agent's tendency to overfit early data distribution and lose the ability to learn from new data distribution effectively. Previous studies \cite{chen2021randomized,li2022efficient,nikishin2022primacy} have shown that such primacy bias due to a high UTD ratio can seriously damage the agent's performance in model-free setting. However, we notice that in model-based reinforcement learning algorithms (MBRL), where a world model is utilized, it is common to use a high UTD ratio for updates to improve sample efficiency without causing significant harm to the agent's performance. For example, MBPO \cite{janner2019trust} employs a high UTD ratio of at least 20 to update the agent, which is enough for MFRL algorithms like SAC \cite{haarnoja2018soft} and TD3 \cite{fujimoto2018addressing} to fail \cite{li2022efficient}. However, MBPO has demonstrated superior performance on many continuous control tasks. This phenomenon raises the following questions: \textit{Does the primacy bias also exist in model-based setting? Does the primacy bias in MBRL differ from its form in model-free scenario if it does exist?}

\begin{figure}[t]
\centering
\includegraphics[width=\columnwidth]{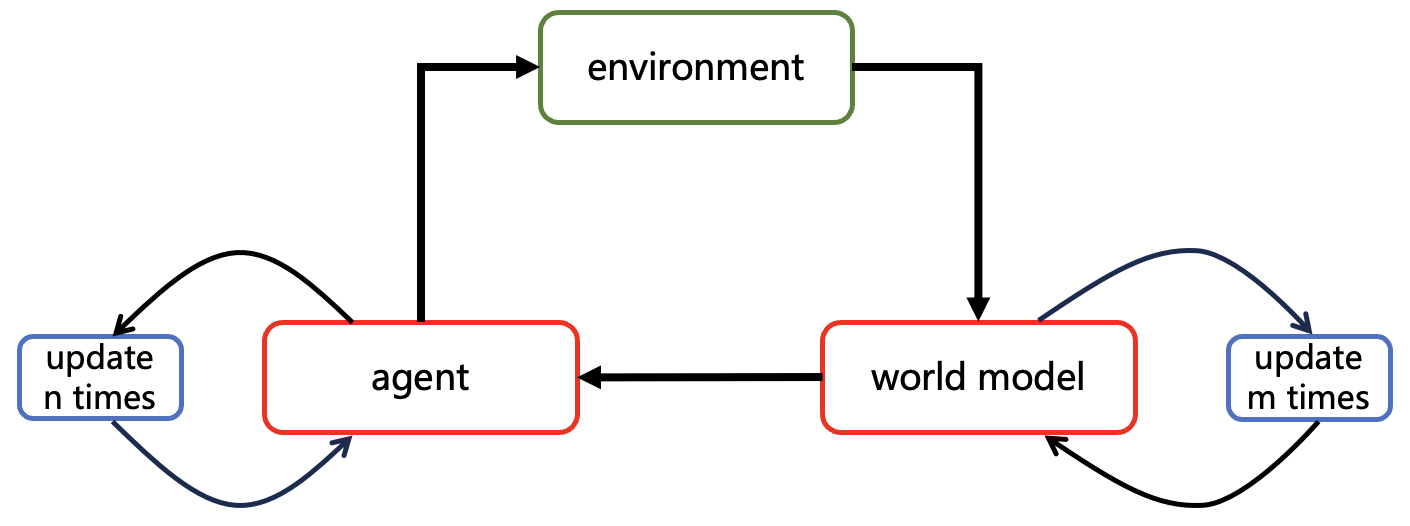}
\caption{The basic flowchart of MBRL algorithms. For each data sampled from the environment, the world model is updated $m$ times. The agent then is trained on data generated by the world model $n$ times. In practice, $n$ is a positive integer, and $m$ is a fraction, which means that the world model is updated every multiple environment steps.} 
\label{mbrl}
\end{figure}

In this paper, we focus on exploring the primacy bias in MBRL, which has been neglected in prior studies. We start by assuming that there are two forms of the primacy bias: \textit{the primacy bias of the agent} and \textit{the primacy bias of the world model}. Next, we apply the parameter resetting technique, a simple yet effective approach to reducing the primacy bias proposed by \cite{nikishin2022primacy}, in MBPO. Surprisingly, instead of improving the agent's performance, we observe a decline in performance compared to the original algorithm. This suggests that parameter resetting seems not to work in model-based setting. To further investigate this phenomenon, we design and conduct extensive experiments. Our findings reveal that the primacy bias still exists in model-based setting. However, it is more closely related to the primacy bias of the world model, which differs from the situation in model-free setting. Building upon this finding, we propose \textit{world model resetting}, which is a modification to the parameter resetting method. This slight modification works in model-based setting, while original parameter resetting does not. To validate the effectiveness of \textit{world model resetting}, we apply it to two MBRL algorithms: MBPO and DreamerV2 \cite{hafner2020mastering}. We conduct extensive experiments on multiple continuous control tasks on MuJoCo \cite{todorov2012mujoco} and DeepMind Control Suite (DMC) \cite{tassa2018deepmind}, as well as discrete control tasks on Atari 100k benchmark \cite{kaiser2019model}. The results show that \textit{world model resetting} can reduce the primacy bias and improve the performance of MBRL algorithms in both continuous and discrete domains. We also compare \textit{world model resetting} with AutoMBPO \cite{lai2021effective}, a parameter tuning method designed specifically for MBPO, to demonstrate the superiority of our method under high UTD ratio conditions. In the final part of our experiments, we examine the key factors that influence the effectiveness of \textit{world model resetting}. This provides a guideline on effectively utilizing our method for better performance.


\begin{figure}[t]
\centering
\includegraphics[width=\columnwidth]{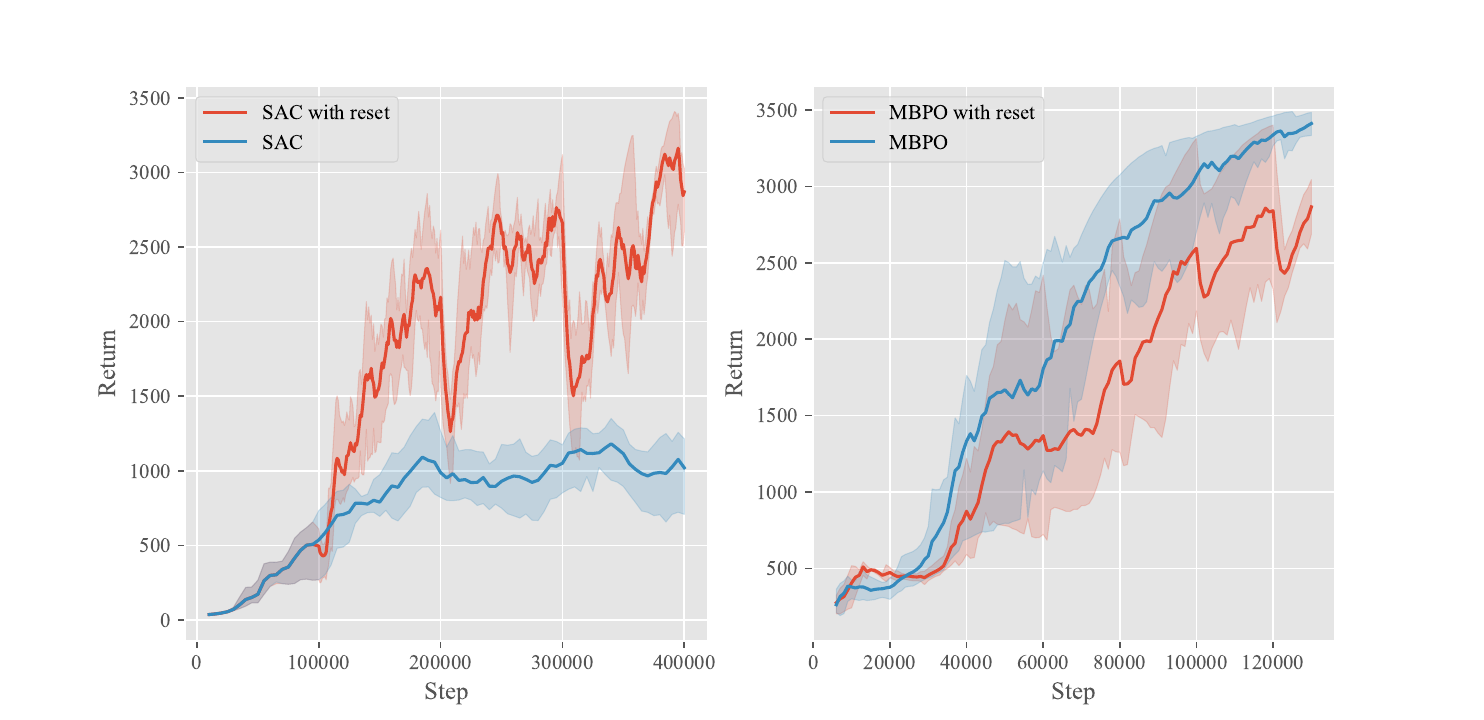}
\caption{\textbf{Left:} Learning curves for Hopper-v2 task of MuJoCo for SAC and SAC with parameter resetting (abbreviated as SAC with reset). ~\textbf{Right:} Learning curves for Hopper-v2 task of MuJoCo for MBPO and MBPO with parameter resetting (abbreviated as MBPO with reset). The solid line represents the average return, the shaded area represents the 95\% confidence interval. The results are evaluated by 5 runs.} \label{reset_compare}
\end{figure}

\section{The Primacy Bias in MBRL}
The main goal of our work is to explore the primacy bias in MBRL. In model-free RL algorithms, the primacy bias typically refers to the agent's tendency to overfit early training data. However, in model-based RL algorithms, we point out that there are two forms of the primacy bias:\\

\noindent \textbf{The Primacy Bias of the agent:} ~ ~ The Q-network and the policy network overfit the early data in the replay buffer and cannot effectively use the data in the later stage of training, resulting in learning a suboptimal policy.\\

\noindent \textbf{The Primacy Bias of the world model:} ~ ~ The world model overfits the data distribution under the early policy and cannot effectively represent the environment dynamics under the evolving policy, failing the algorithm.\\

Unlike model-free RL algorithms, which only suffer from the first form of the primacy bias, the above two forms of primacy bias can appear in MBRL algorithms at the same time, and we assume that either of them may cause the failure of the entire algorithm. In Figure \ref{mbrl}, we present the sources of these two forms of primacy bias. We refer to $m$ as model UTD ratio and $n$ as agent UTD ratio. A high model UTD ratio may lead to overfitting of the world model, which magnifies the primacy bias of the world model. Similarly, a high agent UTD ratio can potentially lead to overfitting of the agent, which stretches the primacy bias of the agent. 

In the following part of this section, we first apply the parameter resetting technique to MBRL but observe that it can't improve the algorithm's performance. Instead, it results in performance degradation. Based on this observation, we empirically investigate the impact of the two forms of the primacy bias on the performance of MBRL algorithms. As a conclusion, we identify the main form of the primacy bias in model-based setting.

\begin{figure}[t]
\centering
\includegraphics[width=\columnwidth]{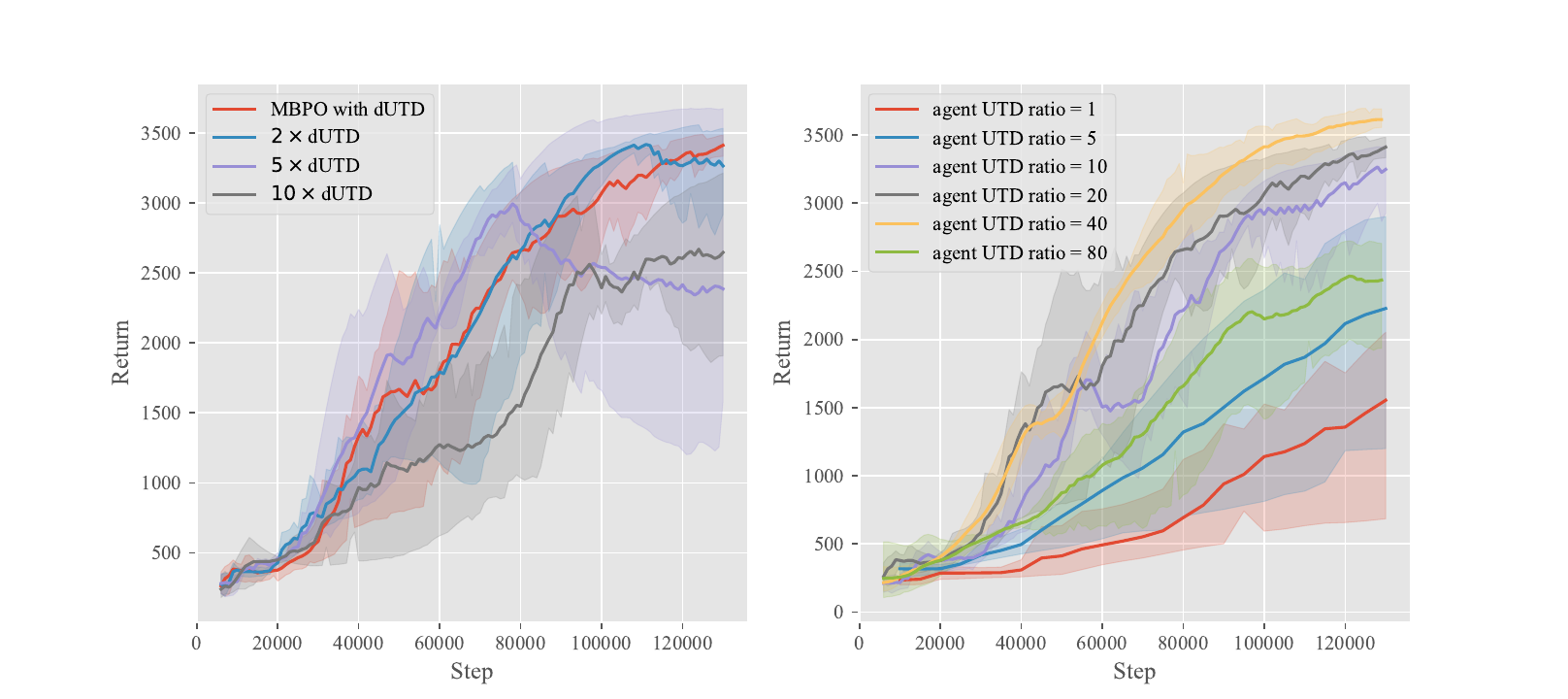}
\caption{\textbf{Left:} Learning curves for Hopper-v2 of MuJoCo for MBPO with different model UTD ratio. ~\textbf{Right:} Learning curves for Hopper-v2 of MuJoCo for MBPO with different agent UTD ratio. The solid line represents the average return, the shaded area represents the 95\% confidence interval. The results are evaluated by 5 runs.} \label{utd_compare}
\end{figure}

\subsection{Parameter resetting doesn't improve the performance}
Resetting the agent's parameters periodically has been shown a simple yet powerful technique to eliminate the primacy bias in model-free scenario, thus improving the algorithm's performance, especially in high agent UTD ratio scenario. But in MBRL setting, can parameter resetting still be effective? 

We apply parameter resetting to model-free RL algorithm SAC and model-based RL algorithm MBPO to answer that question. We experiment on Hopper-v2 task on MuJoCo. To make the effect of parameter resetting as significant as possible, we refer to the experiment settings suggested in \cite{nikishin2022primacy}. For SAC, we set the agent UTD ratio to 32, set the reset interval to $1\times 10^5$ environment steps, and reset the entire Q-network (including target Q-network) and policy network. For MBPO, the optimization algorithm used is also SAC; we set the reset interval to $2\times 10^4$ environment steps, set the agent UTD ratio to 32 (the default agent UTD ratio is 20), and the other hyper-parameters are consistently with the default setting. We train SAC for a total of 400k environment steps and MBPO for 130k environment steps. Figure \ref{reset_compare} shows our experimental results. We can observe that SAC with a high agent UTD ratio has poor performance, but the algorithm performance significantly improves with parameter resetting, consistent with previous studies. However, for MBPO, a high agent UTD ratio does not lead to worse performance. Applying parameter resetting actually harms the performance instead of improving it. To investigate this phenomenon, we conduct further experiments as follows.

\subsection{High model UTD ratio hurts the performance}
The phenomenon observed in Figure \ref{reset_compare} seems quite perplexing. Why is parameter resetting effective in model-free scenarios but not in model-based setting? Does this imply the absence of the primacy bias in MBRL?

\begin{figure*}[t]
\centering
\includegraphics[width=0.9\textwidth]{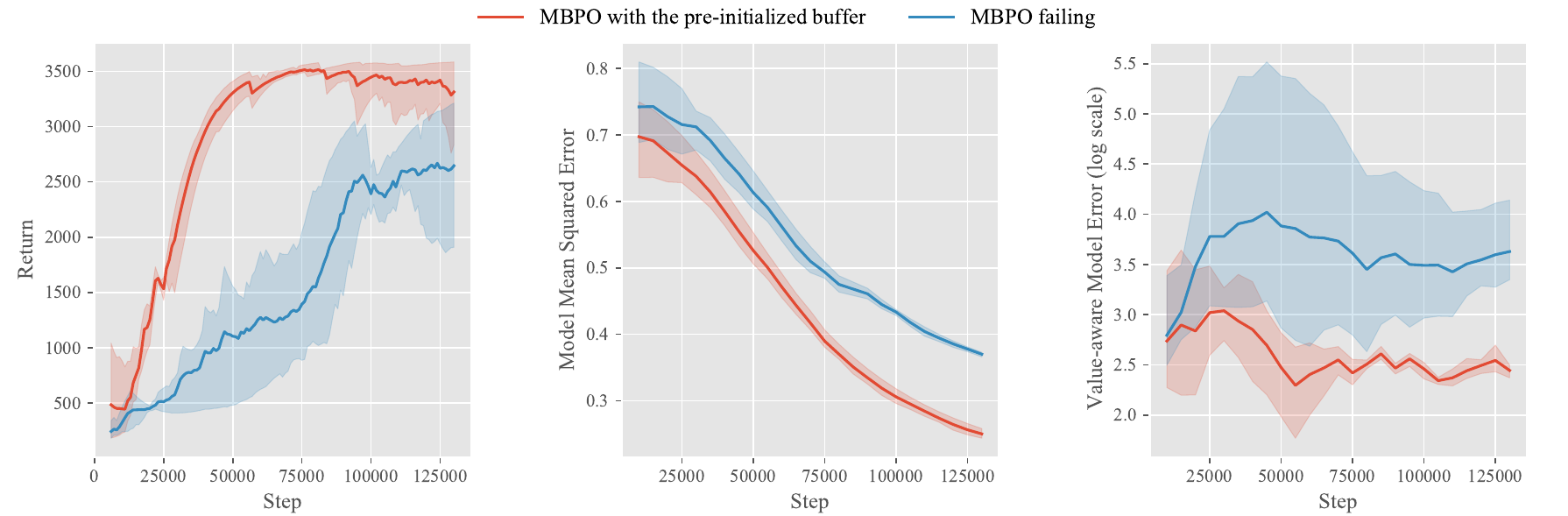}
\caption{ \textbf{Left:} Performance comparison between MBPO and MBPO with a pre-initialized buffer on Hopper-v2 task. ~\textbf{Middle:} Model mean squared error comparison between MBPO and MBPO with a pre-initialized buffer.  ~\textbf{Right:} Value-aware model error comparison between MBPO and MBPO with a pre-initialized buffer. All the results are evaluated by 5 runs. } \label{error}
\end{figure*}

To answer these questions, we conduct further experiments. To investigate the existence of the two forms of primacy bias and their impact on the performance of MBRL algorithms, we conduct experiments still using MBPO algorithm on Hopper-v2 task of MuJoCo. First, to investigate the effect of the primacy bias of the world model on algorithm's performance, we keep the other hyperparameters at their default settings and only modify the model UTD ratio; we observe how this change affects the algorithm's performance. The default model UTD ratio in the original MBPO paper is $\frac{1}{250}$, which means the world model is updated every 250 environment steps. We denote the default model UTD ratio as dUTD for convenience. In addition to dUTD, we set the model UTD ratio to $2\times$ dUTD, $5\times$ dUTD, and $10\times$ dUTD for four experimental groups. We train MBPO and plot the learning curves in Figure \ref{utd_compare} Left. We can observe that when the model UTD ratio doubles, algorithm's performance barely differs. However, when model UTD ratio becomes 5 times and 10 times, there is a significant decrease in the performance of MBPO. This indicates the existence of the primacy bias of the world model, and it does not require a high model UTD ratio (even at 10 times the default model UTD ratio, which is only $\frac{1}{25}$ ) for the primacy bias of the world model to affect algorithm's performance significantly. Similarly, to investigate whether the primacy bias of the agent affects algorithm's performance, we change the agent UTD ratio while keeping other hyperparameters default and conduct sets of experiments. We vary the agent UTD ratio to 1, 5, 10, 20, 40, 80, conducting a total of six groups of experiments. We plot the learning curves in Figure \ref{utd_compare} Right. We observe that the performance of MBPO improves with the increase of the agent UTD ratio, until the ratio reaches 80, after which the performance begins to decline. Compared with SAC, where the algorithm's performance is impaired when the agent UTD ratio is 20, and increasing the model UTD ratio by only 5 times affects the performance of MBPO, this experimental result suggests that in the setting of MBRL, the primacy bias of the agent is not as significant as MFRL, while the primacy bias of the world model is more pronounced.  

The question is, why are the types of the primacy bias different in model-based and model-free setting? We can explain it as follows: Due to the agent in MBRL being trained with data generated by the world model, a couple of factors are at play. Firstly, the number of samples the model generates is typically much larger than the number of samples gathered from the real environment. This abundance of data reduces the likelihood of overfitting. Secondly, the samples generated by the model contain noise, which can be seen as a form of data augmentation. This means that even if the agent UTD ratio is high, it is unlikely to suffer from overfitting, as is often observed in model-free setting. In fact, if the agent UTD ratio is not high enough, underfitting can occur, leading to a decline in performance. On the other hand, when the world model is trained with samples from the real environment, overfitting becomes more likely.

In summary, we draw an important conclusion that: \textbf{in the setting of MBRL, the primacy bias is more towards the primacy bias of the world model, rather than the primacy bias of the agent.}

\subsection{The data in the replay buffer is enough for learning the world model}
Once the world model overfits the early data, even if there is high-quality data in the environment buffer (we call it $D_{env}$), the world model cannot learn from that data efficiently. Actually, even though the agent trained with a high UTD ratio updated model performs poorly, the data in $D_{env}$ is usable. Still, the model cannot effectively learn from it. We design experiments to verify this. We save the MBPO agent trained with model UTD ratio of 10. Due to the algorithm's poor performance, this agent's policy is far from the optimal policy. As a result, the collected data in $D_{env}$ is also poor, but does this poor data hinder the learning process of the world model? To answer that question, we train another MBPO algorithm with the same model UTD ratio of 10. The difference is that we use the saved policy to initialize the replay buffer, rather than a random policy. We plot the learning curves in Figure \ref{error} Left. It shows that the learning speed of the agent with the pre-initialized buffer is much faster, and the final return is higher, which means the data in the pre-initialized buffer is suitable for policy learning. To further show the quality of the world model, we also record the change process of Model Mean Squared Error (MMSE) and Value-aware Model Error (VME). MMSE measures the difference between model MDP and true MDP, and VME measures the error between model Bellman operator and true Bellman operator given a Q function. The specific definition and calculation of MMSE and VME can be found in Appendix A. We plot the two kinds of model error in Figure \ref{error} Middle, Right. We can see that both model errors are lower with the pre-initialized buffer, which means the data in the pre-initialized buffer is enough for learning a world model.

\section{Reset the world model instead of the agent}
In the previous section, we verify the existence of the primacy bias in MBRL and show that, the primacy bias of the world model is more pronounced than the primacy bias of the agent.

Based on these conclusions, we can infer that the parameter resetting technique is not ineffective but rather used in the wrong way. We advocate that parameter resetting can still improve the performance of MBRL algorithms; only a minor adjustment is needed: reset the parameters of the world model, instead of the agent. We call this method \textit{world model resetting}, summarized as follows:

\begin{tcolorbox}[colback=blue!10!white,colframe=blue!70!black,title=world model resetting]
    During the MBRL algorithm training process, periodically resetting the parameters of the \st{agent} \textcolor{red}{world model}, and the rest of the training process remains unchanged.
\end{tcolorbox}

\begin{algorithm}[tb]
\caption{MBPO with model reset}
\label{alg:mbporeset}
\begin{algorithmic}[1] 
\STATE \textbf{Require:} total environment steps $N$, agent UTD ratio $U_a$, model UTD ratio $U_m$, reset interval $I_r$, initial policy $\pi_\theta$, predictive model $p_\phi$, environment dataset $\mathcal{D}_{env}$, model dataset $\mathcal{D}_{model}$

\FOR{step from 1 to $N$}
\STATE Take action in environment according to $\pi_\theta$; add to $\mathcal{D}_{env}$
\STATE Sample $s_t$ uniformly from $\mathcal{D}_{env}$
\STATE Perform $k$-step model rollout starting from $s_t$ using policy $\pi_\theta$; add to $\mathcal{D}_{model}$
\FOR{agent step in 1 to $U_a$}
\STATE Update policy $\pi_\theta$ using data from $\mathcal{D}_{env}\cup\mathcal{D}_{model}$: $\theta\leftarrow \theta-\lambda_\pi\nabla_\theta J_\pi(\theta, \mathcal{D}_{env}\cup\mathcal{D}_{model})$
\ENDFOR
\FOR{model step in 1 to $U_m$}
\STATE Train model $p_\phi$ on $\mathcal{D}_{env}$ via maximum likelihood
\ENDFOR
\IF{step \% $I_r$ == 0}
\STATE Reset the world model
\ENDIF
\ENDFOR
\end{algorithmic}
\end{algorithm}

The complete pseudo-code of applying \textit{world model resetting} to MBPO is presented in Algorithm~\ref{alg:mbporeset}.

\section{Experiments}

This section aims to validate the effectiveness of \textit{world model resetting} (abbreviated as model reset). To achieve this, we design and conduct numerous experiments on both continuous and discrete tasks. We aim to answer the following questions: (1) Can model reset improve the performance of MBRL algorithms? (2) Is the world model with model reset more accurate? (3) Can model reset show superiority over parameter tuning methods? (4) What are the key factors that influence the success of model reset?

\subsection{Overall performance of model reset}
The first question is: can model reset improve the performance of MBRL algorithms? To answer that question, we apply model reset to two MBRL algorithms: MBPO and DreamerV2. We choose them because they utilize different types of world model: probabilistic edynamics model and recurrent state-space model (RSSM), respectively. For MBPO, we conduct experiments on four continuous control tasks on MuJoCo. For DreamerV2, we conduct experiments on multiple continuous control tasks on DMC and discrete control tasks on Atari 100k benchmark to verify the effectiveness of model reset across different domains.

\begin{figure}[t]
\centering
\includegraphics[width=0.5\textwidth]{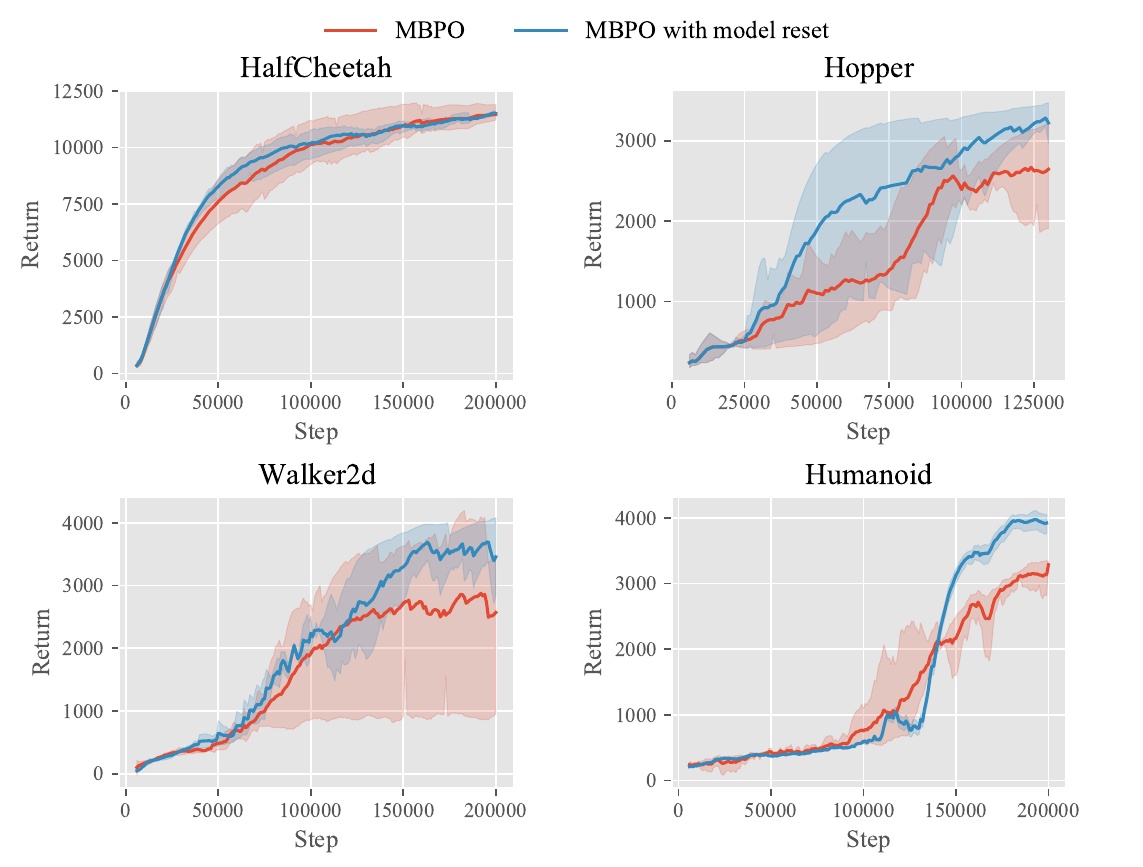}
\caption{Learning curves for MBPO and MBPO with model reset on four tasks of MuJoCo. Model UTD ratio is 10. The results are evaluated by 5 runs.} \label{mujoco_compare}
\end{figure}

\begin{table}
\caption{Point estimates and 95\% confidence interval for the performance of DreamerV2 and DreamerV2 with model reset on 20 tasks on DMC. The results are evaluated over 5 seeds.}
\begin{tabularx}{\columnwidth}{X X X X}
     \toprule
     Method & IQM & Median & Mean \\
     \midrule
     DreamerV2 & 762(716,796) &772(732,814) & 765(722,801)\\ \\
     DreamerV2 + reset & \textbf{783}(732,805) & \textbf{791}(740,819) & \textbf{786}(734,810) \\
     \bottomrule
\end{tabularx}

\label{DMC}
\end{table}

For the world model setting of MBPO and DreamerV2, we use the networks and default hyperparameters as specified in the original paper. For MBPO, the world model is an ensemble probabilistic model consisting of 7 models, each with 4 hidden layers. We reset all ensemble models and only reset the last 2 hidden layers every $2\times10^4$ environment steps. For DreamerV2, the world model consists of an image encoder and a RSSM, and the RSSM consists of a recurrent model, a representation model, and a transition predictor. Among these components, the transition predictor is directly used for behavior learning. Therefore, we only focus on resetting the transition predictor. Due to the complexity of RSSM, we opt for a more conservative Exponential Moving Average (EMA) approach to reset the parameters, which means $\phi_{t+1}=(1-\alpha)\phi_{t}+\alpha\phi_{random}$. $\phi$ represents the parameters of the transition predictor, and we set $\alpha$ to 0.8. The transition predictor consists of one hidden layer, and we reset this hidden layer every $2\times 10^5$ environment steps.

\begin{figure*}[htb]
\centering
\includegraphics[width=1\textwidth]{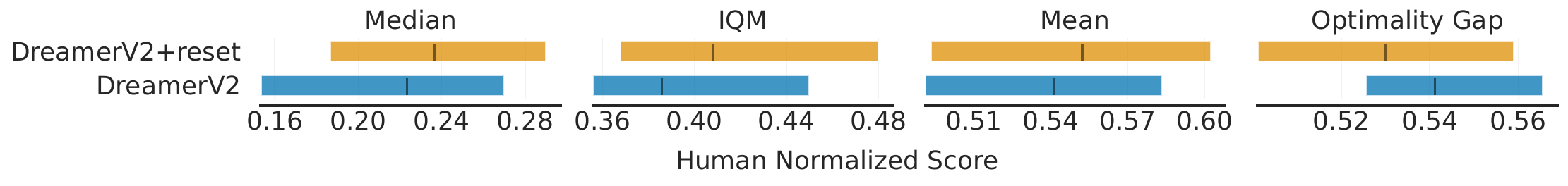}
\caption{Point estimates and 95\% confidence interval for the performance of DreamerV2 and DreamerV2 with model reset on Atari 100k. The metric is suggested by \cite{agarwal2021deep}. The results are evaluated over 5 seeds.} \label{atari}
\end{figure*}

\subsection{Model reset enhances model accuracy}

\begin{figure*}[htb]
\centering
\includegraphics[width=1\textwidth]{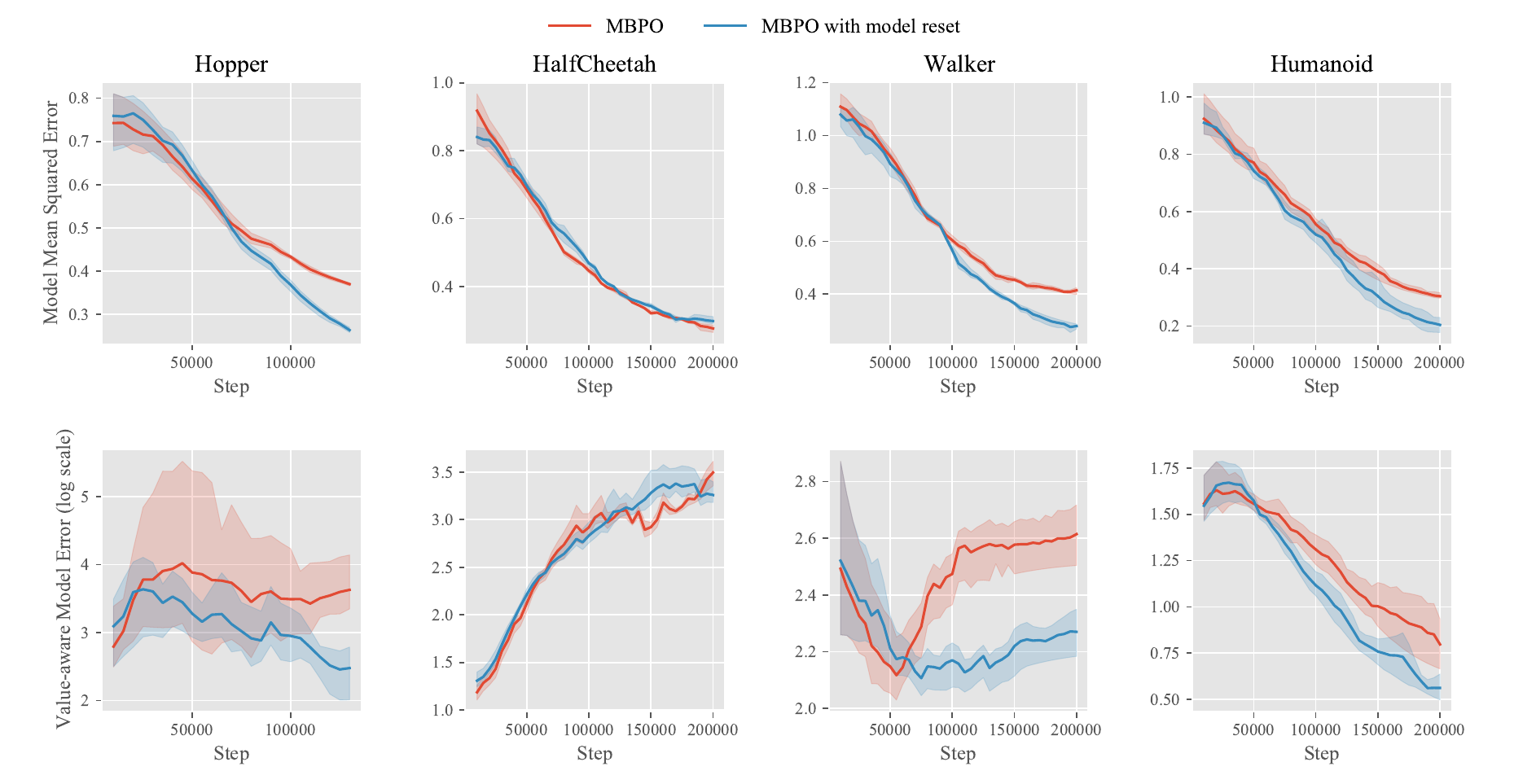}
\caption{Model mean squared error and Value-aware model error comparison between MBPO and MBPO with model reset. The experiments are conducted on four tasks on MuJoCo. The results are evaluated by 5 runs.} \label{fig:3}
\end{figure*}

We plot our results for MBPO with model UTD ratio of 10 on MuJoCo tasks in Figure \ref{mujoco_compare}. Except for HalfCheetah-v2 task, which appears to be unaffected by the primacy bias, the other three tasks significantly improve algorithm's performance with model reset. As for DreamerV2, we show the aggregated results on Atari 100k benchmark and DMC in Figure \ref{atari} and Table \ref{DMC}. We report \textbf{the best results over different model UTD ratios} for methods with and without model reset. The results indicate that the algorithm's performance has improved in continuous and discrete control tasks with model reset.\footnote{One might argue that the results in Table \ref{DMC} show similar performance with and without reset. We point out that the results here are the best results over different model UTD ratios, and the effect of model reset is more obvious for a fixed high model UTD ratio, which we show in Appendix C.1.} We include more detailed experimental results in Appendix C.1.

The next question is: is the world model with model reset more accurate at collected data points? This question is crucial because it determines whether model reset improves algorithm's performance by enhancing the model's accuracy, rather than other potential factors. To answer that question, we conduct experiments to evaluate the world model in MBPO during the training process. We do not choose DreamerV2 because its world model component is too complex and relies on temporal states, making it difficult to evaluate the model.

\begin{figure}[tb]
\centering
\includegraphics[width=\columnwidth]{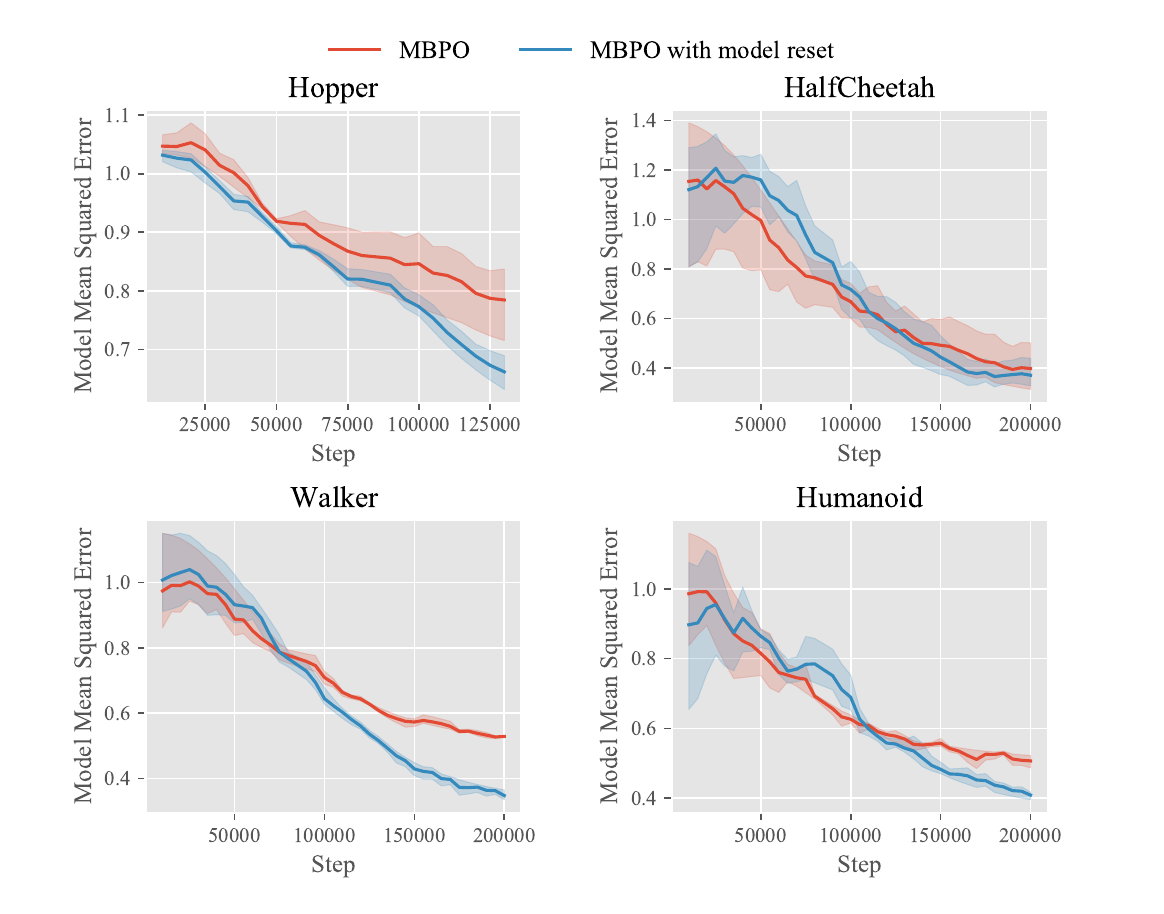}
\caption{Model mean squared error over the data in reserved SAC replay buffer. The experiments are conducted on four tasks of MuJoCo. The results are evaluated by 5 runs.} \label{fig:4}
\end{figure}

We first observe the changes in model error (including model mean square error and value-aware model error) during the training process of MBPO and MBPO with reset. The results are presented in Figure \ref{fig:3}. The results indicate that in the later stage of training, the model with model reset can better fit the data in the current buffer. Periodic reset allows the model to learn from new data distribution effectively. In addition to the ability to fit the current data, we also want to know the model's ability to predict the environment dynamics under the near-optimal policy. This can reflect whether the policy has the potential to reach the optimal policy. Therefore, we additionally train a SAC agent on four tasks of MuJoCo for 1 million steps. We keep the replay buffer at the end of the training. When training MBPO, we use the reserved replay buffer to calculate model mean square error. We present the curves in Figure \ref{fig:4}. The results show that the model's ability to predict the dynamics under near-optimal policy has improved with model reset. This suggests that the algorithm has greater potential to reach the optimal policy. In fact, a good model does not need to be accurate on every data point but rather should be accurate on the data points that the policy frequently visits. Model reset can be seen as a forgetting mechanism that allows the model to focus more on the data distribution under the current policy, thereby obtaining a better model.

\subsection{Comparison with hyperparameter tuning methods}

In the previous sections, we empirically show that model reset can mitigate the primacy bias for MBRL algorithms and improve the algorithms' performance. In this part, we aim to answer the question:  \textbf{under high model UTD ratio conditions where the primacy bias is pronounced}, whether model reset can show superiority over hyperparameter tuning methods, i.e., tuning real data ratio or rollout length? 

As we discuss in Section 2.2, the primacy bias of the world model can be pronounced under a high model UTD ratio, thus harming the performance of MBRL algorithms. A natural idea is to automatically tune model UTD ratio to mitigate the primacy bias. Previous studies follow this idea and propose hyperparameter tuning methods, i.e., \cite{dorka2023dynamic} utilizes validation sets for dynamically adjusting the model UTD ratio, \cite{ji2022update} introduces an event-triggered mechanism to determine when to update the model. Although adjusting model UTD ratio is an effective way to mitigate the primacy bias, it comes at a risk of sacrificing sample efficiency. Therefore, we set model UTD ratio to a large fixed value, intending to compare the effects of model reset and tuning hyperparameters other than model UTD ratio on the performance of MBRL algorithms.

We then introduce our experimental setup. We choose MBPO as our base MBRL algorithm and set the model UTD ratio to $10\times$dUTD, which is high enough to show primacy bias. As for the hyperparameter tuning method, we choose AutoMBPO~\cite{lai2021effective}, a hyperparameter adjusting method which introduces a parametric hyper-controller to sequentially select the value of hyperparameters to maximize the performance of MBPO. The hyper-controller dynamically adjusts hyperparameters including real data ratio, model UTD ratio, agent UTD ratio and rollout length during the training process of MBPO. Since we fix the model UTD ratio, we only adjust real data ratio, agent UTD ratio and rollout length. For model reset, we follow the setup in Section 4.1, resetting all ensemble models and resetting only the last 2 hidden layers every $2\times10^4$ steps. 

We conduct experiments on four tasks of MuJoCo, comparing the performance of AutoMBPO and MBPO with model reset. The experimental results are shown in Figure~\ref{fig:15}. We can observe that model reset can incur higher overall performance improvement than AutoMBPO under the model UTD ratio of 10. Moreover, the implementation of AutoMBPO is complex since it contains an additional hyper-controller while model reset only needs a few lines of code to change. These demonstrate the advantage of model reset under high model UTD ratio conditions.

\begin{figure}[t]
\centering
\includegraphics[width=\columnwidth]{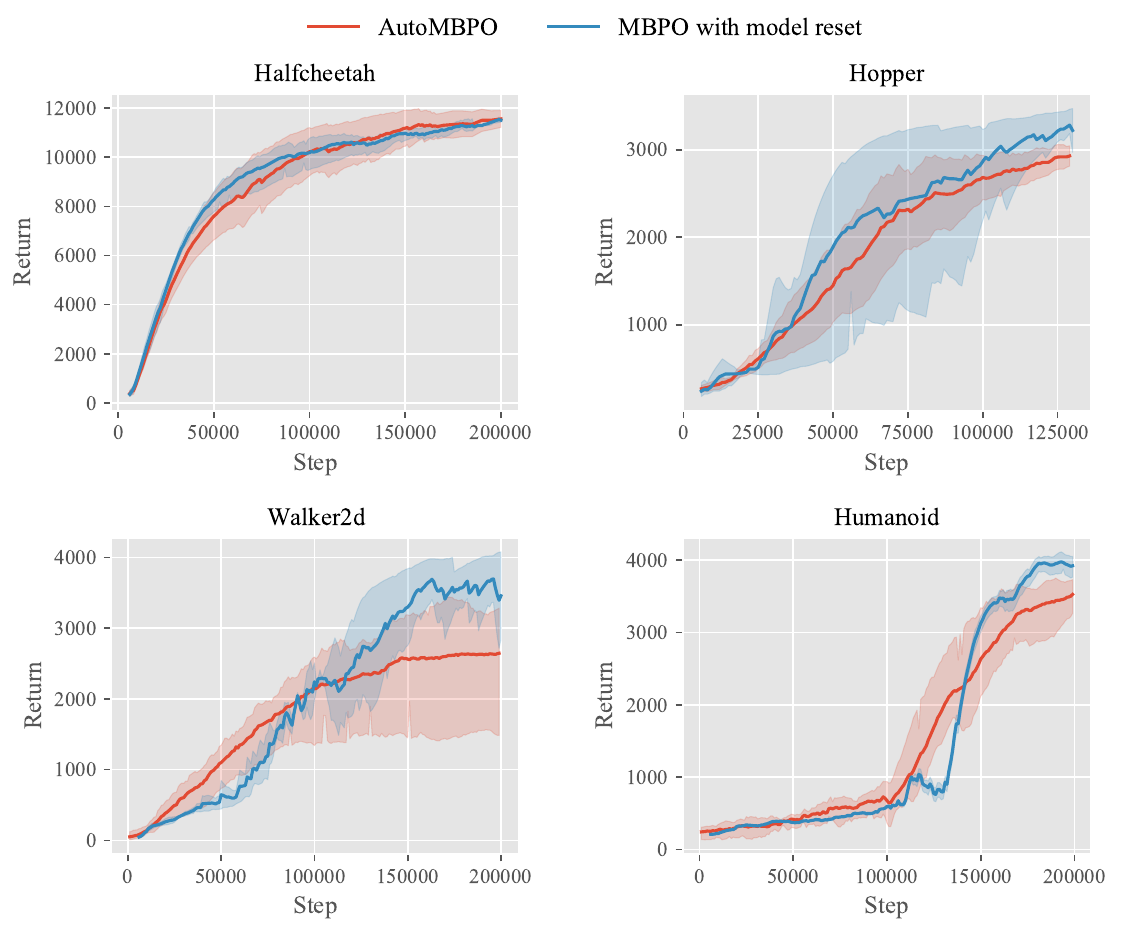}
\caption{Learning curves for AutoMBPO and MBPO with model reset on four tasks of MuJoCo. Model UTD ratio is 10. The results are evaluated by 5 runs.} \label{fig:15}
\end{figure}

\subsection{The key factors influencing model reset}
In this section, we aim to investigate the key factors that influence the effectiveness of model reset and maximize the use of model reset to improve the performance of MBRL algorithms. Due to space limit, we only present our main points and conclusions here. The detailed results can be found in Appendix C.\\

\noindent \textbf{Model UTD ratio} ~ Model UTD ratio determines the extent to which the model utilizes collected data. A high model UTD ratio implies that the model is prone to fit the current data, potentially increasing the risk of overfitting. In our experiments, we observe the model UTD ratio significantly impacts the effect of model reset. We vary the model UTD ratio of MBPO and DreamerV2 and observe how model reset affects the degree of performance improvement. For MBPO, we conduct experiments on four tasks on MuJoCo. We vary the model UTD ratio to be $2\times$ dUTD, $5\times$ dUTD, and $10\times$ dUTD. For DreamerV2, we conduct experiments on 20 tasks on DMC and 26 tasks on Atari 100k. For DMC, We change the default model UTD ratio ($\frac{1}{5}$) to 5 times and 10 times. For Atari 100k, We change the default model UTD ratio ($\frac{1}{16}$) to 4 times and 8 times. Experimental results indicate that, as the model UTD ratio increases, the level of overfitting of the model also increases. Therefore, the degree of performance improvement from model reset is also higher.\\



\noindent \textbf{Model ensemble size} ~ In MBRL algorithms, such as MBPO, the world model can be an ensemble model. The ensemble model utilizes ensemble learning to reduce the variance of prediction and thus mitigate the risk of overfitting to some extent. Therefore, model ensemble size is also one of the potential influencing factors. We conduct experiments using MBPO. We vary the ensemble size to 1, 3, 5, and 7 and apply model reset. The results indicate that as model ensemble size increases, the effectiveness of model reset diminishes.\\

\noindent \textbf{When and how to reset the model}
~ The last factor we examine is how to conduct model reset, such as how often to perform a reset and which part of the network to reset. These factors may have a significant impact on the effectiveness of model reset. Through this investigation, we hope to provide insights and a guideline on performing model reset effectively.

In our experiments, if the model has multiple hidden layers, resetting only the last few layers yields the best results compared to resetting the first few hidden layers or randomly resetting hidden layers. We conjecture that the first few layers can extract the general features of the task \cite{ota2020can}, allowing the world model to adapt to the new data distribution after a reset quickly. Therefore, retaining the first few hidden layers when performing reset is preferable. We also find that, for ensemble model, resetting the majority or all ensemble members has a better effect than resetting only one or few of them. Regarding when to perform reset, we find that it depends on the specific task and model UTD ratio. Generally, if the model UTD ratio is higher, the reset frequency should also be higher.

\section{Related Work}
\textbf{The primacy bias in reinforcement learning}~The primacy bias is originally a concept in cognitive science \cite{marshall1972effects}, which means first impressions influence human consciousness and people tend to pay less attention to subsequent impressions. \cite{nikishin2022primacy} first propose that in deep reinforcement learning, the primacy bias also exists, and is related to the agent's overfitting of early data. 
In fact, the phenomenon of overfitting \cite{zhang2018study,zhang2018dissection} in deep reinforcement learning has been studied a lot. \cite{sutton2018reinforcement} point out that when using the Q function for generalization, there might be overfitting; \cite{lyle2022understanding} show the approximator will lose the ability to generalize during training in off-policy RL; \cite{cetin2022stabilizing} show the "self-overfitting" issue exists in image-based RL with an image decoder. It is worth mentioning that overfitting and primacy bias in RL cannot be equated. For example, overfitting can also occur in offline setting \cite{fujimoto2019off,kumar2020conservative}, but since the dataset is static, it cannot be called the primacy bias. The primacy bias often occurs in high UTD ratio scheme \cite{van2019use,fedus2020revisiting}, causing the failure of the algorithm. Previous works attempt to handle this in various ways, such as randomized ensemble \cite{chen2021randomized}, network normalization \cite{hiraoka2021dropout,smith2022walk}, replay ratio scaling \cite{d2022sample}, and so on. \cite{nikishin2022primacy} propose a simple method called parameter resetting. By periodically resetting the parameters of the agent, it can effectively deal with the scheme with a high UTD ratio. These methods are mainly about model-free setting, while we further focus on model-based setting. \\

\noindent \textbf{Overfitting in model-based RL} ~ In model-based setting, most previous work focuses on handling the issue that the agent overfits an inaccurate world model \cite{jiang2015dependence,arumugam2018mitigating}. Commonly used methods are uncertainty estimation \cite{chua2018deep}, and policy regularization \cite{zheng2023model}. We need to mention that the overfitting here is different from the primacy bias we discuss above, because it is caused by an inaccurate model, and we focus on the overfitting caused by a high UTD ratio. There are also studies focusing on the overfitting of the world model itself. Standard solutions include using different model structures, such as Bayesian neural networks \cite{okada2020variational}, and supervised learning methods, such as spectral regularization \cite{gogianu2021spectral} and dropout \cite{srivastava2014dropout}. Some recent studies aim to mitigate model overfitting by adjusting model UTD ratio, i.e., \cite{dorka2023dynamic} proposes to dynamically adjust the model UTD ratio by utilizing validation sets. However, different network structures may bring computational burdens, and adjusting UTD ratio may sacrifice sample efficiency. \textit{World model resetting} proposed by us neither changes the network structure nor adjusts UTD ratio, but can be seen as a way to make the world model pay more attention to current transitions, thereby alleviating overfitting. However, previous works do not explore the primary and secondary relationship between agent overfitting and model overfitting, which is the focus of our work.\\

\noindent \textbf{Model adaptation} ~ Our work is also related to model adaptation. In model-based RL, the purpose of model adaptation is to enable the world model to adapt when encountering new data or new tasks quickly. Meta learning \cite{thrun1998lifelong,schweighofer2003meta,khetarpal2022towards} is often used for model adaptation \cite{nagabandi2018learning,lee2020context,guo2022relational}. These methods use a prior model to help the agent quickly adapt to new tasks. Other methods include system identification \cite{kumar2021rma}, curious replay \cite{kauvar2023curious}, drifting Gaussian process \cite{meier2016drifting}. Our work differs from these methods because we focus on model learning in a fixed environment, while they are mainly about a changing environment. PDML \cite{wang2023live} learns a model which can adapt to the evolving policy in the same environment. However, PDML guides the model to pay more attention to the current transitions mainly by re-weighting the transition distribution in the replay buffer, while we consider it from the perspective of the primacy bias, allowing the world model to quickly adapt to the current policy by \textit{world model resetting}, which is simpler to apply. 

\section{Limitations}

Our work has several limitations. One is that model reset is only effective when the primacy bias is significant (such as, in high model UTD ratio scenario). Another limitation is that applying resetting technique to complex network structures (such as Transformers \cite{vaswani2017attention}.) is challenging. However, addressing these issues is beyond the scope of this work. For future work, we can focus on exploring other more effective methods to reduce the primacy bias, which 
 we believe is an interesting topic.

\section{Conclusion}
In this paper, we focus on the primacy bias in MBRL. We first assume that the primacy bias is composed of the primacy bias of the agent and the primacy bias of the world model. However, our experiments find parameter resetting is ineffective for MBRL algorithms. In further investigations, we discover that a high agent UTD ratio benefits the algorithm, while a high model UTD ratio hurts the performance. Therefore, we conclude that the primacy bias in MBRL is primarily composed of the primacy bias of the world model. Additionally, we propose \textit{world model resetting}, which works in MBRL setting. 

In the validation experiments of our method, we demonstrate that \textit{world model resetting} can improve the performance of the MBRL algorithms across multiple domains, and the world model with model reset is more accurate. We also compare \textit{world model resetting} with AutoMBPO to illustrate the advantage of our method under high model UTD ratio conditions. Finally, we investigate the key factors that influence the effectiveness of model reset. Our intention is not to propose a totally new and state-of-the-art method for improving MBRL algorithms' performance. Instead, \textit{world model resetting} is simple yet effective, and we hope our work can shed light on addressing the issues of model overfitting and the primacy bias in MBRL. 

\ack This work was supported by the STI 2030-Major Projects under Grant 2021ZD0201404. The authors also would like to thank the anonymous reviewers for their valuable comments for improving the manuscript.



\bibliography{mybibfile}
\onecolumn
\appendix
\section{Additional preliminaries and background} 
\subsection{Omitted background for MBPO and DreamerV2}
In this subsection, We will briefly introduce MBPO and DreamerV2 omitted in the main text.

MBPO is a dyna-style \cite{sutton1991dyna} MBRL algorithm. The training process of MBPO involves alternately updating the dynamics model and optimizing the policy using the collected data from both real and simulated environments. For model training, MBPO uses a probabilistic ensemble model. It consists of $M$ ensemble members: $\hat{P}_\theta = \{\hat{P}_{\theta_1}, \hat{P}_{\theta_2}, \dots, \hat{P}_{\theta_M}\}$. Each member predicts the environmental dynamics $\hat{P}(s^\prime|s,a)$ and outputs the mean and variance of the Gaussian distribution $\hat{P}_{\theta_i}\equiv\mathcal{N}(\mu_{\theta_i}(s,a), \Sigma_{\theta_i}(s,a))$. Each member is trained by maximum log-likelihood~\cite{janner2019trust}: $\mathcal{L}_{\theta_k}=\sum_{t=1}^N\left[\mu_{\theta_k}(s_t, a_t)-s_{t+1}\right]^T\Sigma_{\theta_k}^{-1}(s_t, a_t)\left[\mu_{\theta_k}(s_t, a_t)-s_{t+1}\right]+log det\Sigma_{\theta_k}(s_t, a_t)$. For policy training, MBPO utilizes the model to generate simulated rollouts to train a SAC agent. Moreover, MBPO uses truncated model rollouts to disentangle the task horizon and model horizon, which is an effective way to alleviate model error. 

DreamerV2 is an MBRL algorithm that is capable of handling image inputs. Similar to MBPO, the training process of DreamerV2 also involves training the world model and behavior learning. However, there are significant differences between DreamerV2 and MBPO. The world model of DreamerV2 is a latent dynamics model where the input states are low-dimensional states extracted from high-dimensional images. The world model consists of a Recurrent State-Space Model (RSSM), an image encoder, and predictors. The image encoder encodes the image with CNN \cite{lecun1989backpropagation}, and predictors decode the image, reward and discount factor from the latent state. RSSM comprises a recurrent model, a representation model, and a transition predictor. During the training process of the world model, the input is a sequence of images, and the loss function is the reconstruction loss of the image, reward, and discount predicted by the predictors, plus a KL loss term constraining the gap between prior and posterior distribution. For behavior learning, the RL agent learns behavior purely from the prediction of the world model in an actor-critic scheme. The critic is trained by temporal-difference learning \cite{sutton2018reinforcement}, and the actor is trained with Reinforce gradients and straight-through gradients. We recommend reading the original paper \cite{hafner2020mastering} for readers wanting more details.

\subsection{Omitted description of Model Mean Squared Error and Value-aware Model Error}
Given a world model which predicts the environment dynamics $\hat{\mathcal{P}}(s^\prime|s,a)$ and environment buffer $D_{env}$, Model Mean Squared Error is defined as $\mathbb{E}_{(s,a)\sim D_{env},s^\prime\sim\mathcal{P}(s^\prime|s,a),\hat{s}^\prime\sim\hat{\mathcal{P}}(s^\prime|s,a)}\left[\left\|s^\prime-\hat{s}^\prime\right\|^2\right]$. Given another value function $V(s)$, Value-aware Model Error can be defined as $\mathbb{E}_{(s,a)\sim D_{env},s^\prime\sim\mathcal{P}(s^\prime|s,a),\hat{s}^\prime\sim\hat{\mathcal{P}}(s^\prime|s,a)}\left[\left(V(s^\prime)-V(\hat{s}^\prime)\right)^2\right]$ \cite{farahmand2017value}. Model Mean Squared Error measures the difference between true and predicted environmental dynamics. It is an indicator that reflects the degree of model fitting. Value-aware Model Error measures the difference between true and simulated Bellman operator during the training process of the value function. This indicator reflects the influence of model prediction error on the value function training process. These two indicators can comprehensively reflect the world model's impact on the algorithm's entire training process. 

In MBPO, the world model is an ensemble probabilistic dynamics model. When we calculate the model error, we calculate the average of the model error of all ensemble members. Specifically, given a batch of transitions $
\{s,a,r,s^\prime\}_N$ from $D_{env}$, we calculate the Model Mean Squared Error as $\frac{1}{M}\sum_{i=1}^M\frac{1}{N}\sum_{j=1}^N\left[\left\|s^\prime-\hat{s}^\prime\right\|^2\right]$. When we calculate Value-aware Model Error, since we don't have a value function $V(s)$, we use $Q(s,\pi(s))$ instead. So we calculate Value-aware Model Error as $\frac{1}{M}\sum_{i=1}^M\frac{1}{N}\sum_{j=1}^N\left[\left(Q(s^\prime,\pi(s^\prime))-Q(\hat{s}^\prime, \pi(\hat{s}^\prime))\right)^2\right]$.\\

\section{Experimental Setup}
\subsection{Codebase}
For our codebase, we use the open-source Pytorch version of MBPO \footnote{The codebase we use for MBPO: \\https://github.com/x35f/unstable\_baselines.git} and DreamerV2 \footnote{The codebase we use for DreamerV2: \\https://github.com/vincent-thevenin/DreamerV2-Pytorch.git} implementation. Although it is different from the official codebase provided in the original paper, the codebase we choose reproduces the results in the original paper to the greatest extent.

\subsection{Benchmarks and tasks}
For MBPO, We conduct all the experiments on four continuous control tasks of MuJoCo. For DreamerV2, we conduct experiments on 20 tasks of DeepMind Control Suite and 26 tasks of Atari 100k benchmark. Table \ref{tab:2} lists the four tasks of MuJoCo for testing MBPO, Table \ref{tab:3} and Table \ref{tab:4} list the tasks on DeepMind Control Suite and Atari 100k benchmark for testing DreamerV2. We also report the number of training steps for each task.

\begin{minipage}{\columnwidth}

\begin{minipage}[t]{0.31\columnwidth}

\makeatletter\def\@captype{table}

\begin{tabular}{cc}
    \toprule
    Task  & Steps \\
    \midrule
    HalfCheetah-v2 & 200k \\
    Hopper-v2 & 130k  \\
    Walker2d-v2 & 200k  \\
    Humanoid-v2 & 200k \\
    \bottomrule
\end{tabular}
\caption{4 Tasks of MuJoCo for MBPO and training steps of each task.}
\label{tab:2}
\end{minipage}
\begin{minipage}[t]{0.31\columnwidth}

\makeatletter\def\@captype{table}

\begin{tabular}{cc}
    \toprule
    Task   & Steps  \\
    \midrule
    Acrobot Swingup & $2\times10^6$   \\
    Cartpole Balance & $1\times10^6$  \\
    Cartpole Balance Sparse & $1\times10^6$ \\
    Cartpole Swingup & $1\times10^6$ \\
    Cartpole Swingup Sparse& $1\times10^6$\\
    Cheetah Run & $2\times10^6$\\
    Cup Catch & $1\times10^6$\\
    Finger Spin & $1\times10^6$\\
    Finger Turn Easy & $2\times10^6$\\
    Finger Turn Hart & $2\times10^6$\\
    Hopper Hop & $2\times10^6$\\
    Hopper Stand & $1\times10^6$\\
    Pendulum Swingup & $1\times10^6$\\
    Quadruped Run & $2\times10^6$ \\
    Quadruped Walk & $2\times10^6$\\
    Reacher Easy & $1\times10^6$\\
    Reacher Hard & $2\times10^6$\\
    Walker Run & $2\times10^6$\\
    Walker Stand & $1\times10^6$\\
    Walker Walk & $2\times10^6$\\
    \bottomrule
\end{tabular}
\caption{20 Tasks of DMC for DreamerV2 and training steps of each task.}
\label{tab:3}
\end{minipage}
\begin{minipage}[t]{0.31\columnwidth}

\makeatletter\def\@captype{table}

\begin{tabular}{cc}
    \toprule
    Task     & Steps  \\
    \midrule
    Alien & $4\times10^5$   \\
    Amidar & $4\times10^5$  \\
    Assault & $4\times10^5$ \\
    Asterix & $4\times10^5$ \\
    Bank Heist & $4\times10^5$\\
    Battle Zone & $4\times10^5$\\
    Boxing & $4\times10^5$\\
    Breakout & $4\times10^5$\\
    Chopper Command & $4\times10^5$\\
    Crazy Climber & $4\times10^5$\\
    Demon Attack & $4\times10^5$\\
    Freeway & $4\times10^5$\\
    Frostbite & $4\times10^5$\\
    Gopher & $4\times10^5$ \\
    Hero & $4\times10^5$\\
    Jamesbond & $4\times10^5$\\
    Kangaroo & $4\times10^5$\\
    Krull & $4\times10^5$\\
    Kung Fu Master & $4\times10^5$\\
    Ms Pacman & $4\times10^5$ \\
    Pong & $4\times10^5$\\
    Private Eye & $4\times10^5$\\
    Qbert & $4\times10^5$\\
    Road Runner & $4\times10^5$\\
    Seaquest & $4\times10^5$\\
    Up N Down & $4\times10^5$\\
    \bottomrule
\end{tabular}
\caption{26 tasks of Atari 100k for DreamerV2 and training steps of each task.}
\label{tab:4}
\end{minipage}
\end{minipage}

\subsection{Hyperparameters}
We list the primary hyperparameters in our experiments here. Table \ref{tab:6} lists the default model UTD ratio and agent UTD ratio of MBPO on each task of MuJoCo. In our experiments, we change these two hyperparameters to explore the primacy bias. Table \ref{tab:7} lists other MBPO hyperparameter setting that are the same across four tasks on MuJoCo. Among them, we only change model ensemble size for the need of investigation, and keep the other hyperparameters unchanged. Table \ref{tab:8} lists the default hyperparameters for DreamerV2 on Atari 100k and DMC, respectively.

\begin{table}
    \centering
    \caption{Default model UTD ratio and agent UTD ratio of each task for MBPO.}
    \begin{tabular}{c c c}
        \toprule
        Task & model UTD ratio & agent UTD ratio \\
        \midrule
        HalfCheetah-v2 & $\frac{1}{250}$ & 40 \\
        Hopper-v2 & $\frac{1}{250}$ & 20 \\
        Walker2d-v2 & $\frac{1}{250}$ & 20 \\
        Humanoid-v2 & $\frac{1}{250}$ & 20\\
        \bottomrule
    \end{tabular}

    \label{tab:6}
\end{table}

\begin{table}
    \centering
    \caption{Other default hyperparameters in our experiment for MBPO.}
    \begin{tabular}{c c}
        \toprule
         Hyperparameters & value  \\
         \midrule
         Model ensemble size & 7\\
         Model hidden layer & (400,400,400,400)\\
         Actor and critic hidden layer & (256,256,256)\\
         Batch size & 256\\
         Activation function & ReLU\\
         Learning rate & 3$\times10^{-4}$\\
         Optimizer & Adam\\
         \bottomrule
    \end{tabular}

    \label{tab:7}
\end{table}

\begin{table}
    \centering
    \caption{Default hyperparameters for DreamerV2 on Atari 100k and DMC.}
    \begin{tabular}{c c c}
        \toprule
        \diagbox{Hyper}{Value}{Domain}
         & Atari & DMC \\
         \midrule
         Model UTD ratio & $\frac{1}{16}$ & $\frac{1}{5}$\\
         Discrete latent dimensions & \multicolumn{2}{c}{32}\\
         Discrete latent classes & \multicolumn{2}{c}{32}\\
         RSSM layer & \multicolumn{2}{c}{(600,)}\\
         Batch size & \multicolumn{2}{c}{50}\\
         World model learning rate & \multicolumn{2}{c}{$2\times10^{-4}$}\\
         Actor learning rate & \multicolumn{2}{c}{$4\times10^{-5}$}\\
         Critic learning rate & \multicolumn{2}{c}{$1\times10^{-4}$}\\
         Optimizer & \multicolumn{2}{c}{Adam} \\
         \bottomrule
    \end{tabular}

    \label{tab:8}
\end{table}

\subsection{Compute Infrastructure}
We run all of our experiments and baselines on two compute infrastructures with the same configuration, and we list their information in Table \ref{tab:5}.
\begin{table}
    \centering
    \caption{Compute infrastructure}
    \begin{tabular}{c|c|c}
        \toprule 
        CPU & GPU & Memory \\
        \midrule
         AMD EPYC 7452 & RTX3090 $\times$8 & 288GB \\
         \bottomrule
    \end{tabular}

    \label{tab:5}
\end{table}

\section{Experimental results}
In this section, we will supplement more detailed experimental results not presented in the main text due to space limitation. We will illustrate how the results support our conclusions in the main text.

\subsection{The impact of model UTD ratio}
\begin{figure}
\centering
\includegraphics[width=\columnwidth]{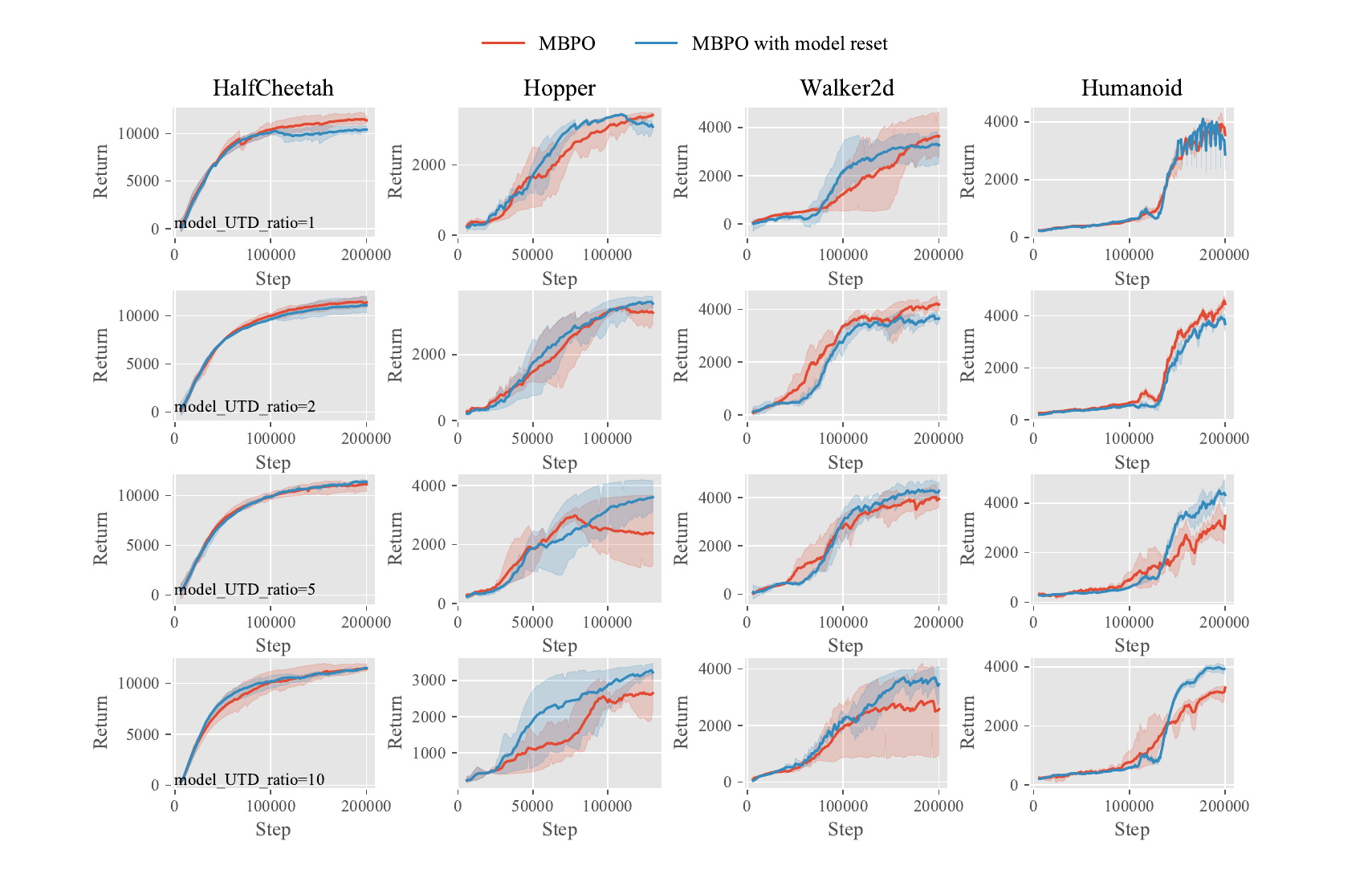}
\caption{Learning curves of MBPO with and without model reset under different model UTD ratios. The solid line represents average return and shaded area represents 95\% confidence interval. All the results are estimated by 5 runs.} \label{fig:9}
\end{figure}

\begin{figure}
\centering
\includegraphics[width=\columnwidth]{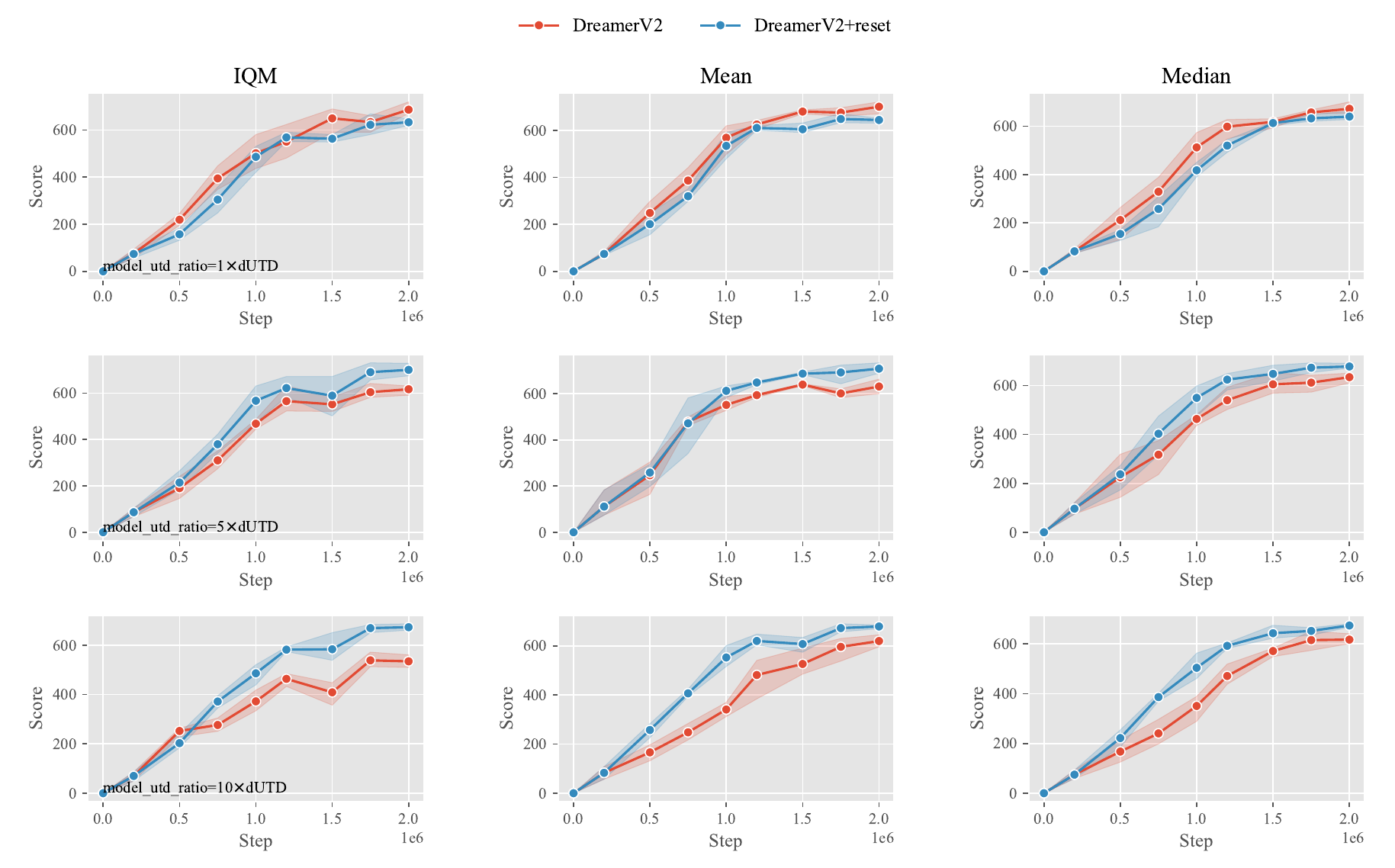}
\caption{Sample efficiency curves of DreamerV2 with and without model reset on DMC under different model UTD ratios. The metrics we record are IQM, Mean and Median score aggregated from 20 tasks on DMC. The results are evaluated by 5 runs.} \label{fig:14}
\end{figure}

This subsection provides more detailed experimental results on the effectiveness of \textit{world model resetting}, as well as a comparison of the performance of the original algorithm and the algorithm with model reset under different model UTD ratios. Figure \ref{fig:9} shows the learning curves of MBPO and MBPO with model reset on four tasks of MuJoCo under different model UTD ratios. For the convenience of analysis, we summarize the corresponding final performance shown in Figure \ref{fig:9} in Table \ref{tab:9}. As shown, when model UTD ratio increases consistently (greater than 5), \textit{world model resetting} boosts the performance of MBPO on three of the four tasks of MuJoCo. For HalfCheetah-v2, model reset has a loss in algorithm performance. Still, for the other three tasks, model reset can reduce the primacy bias's impact under a high UTD ratio scenario, thereby improving algorithm performance. For DreamerV2, Figure \ref{fig:14} and Table \ref{tab:10} show the aggregated results under different model UTD ratios on 20 tasks on DMC. The metrics we record are IQM, Mean, and Median score. Taking the IQM score as an example, increasing model UTD ratio will significantly decrease algorithm performance. Although model reset will damage the performance under a low UTD ratio, it can substantially improve algorithm performance under a high UTD ratio, even exceeding the performance of DreamerV2 under default model UTD ratio. Figure \ref{fig:15} shows the aggregated results on 26 tasks on Atari 100k benchmark. Except for the default UTD ratio and the performance loss caused by model reset, we observe that algorithm performance improves with model reset when model UTD ratio is increased by a factor of 4 and 8. 

All the results indicate that model reset can reduce the primacy bias in MBRL, thus improving algorithm performance. However, the effectiveness of model reset largely depends on the model UTD ratio. If model reset is used in a low model UTD ratio scenario, it may decrease algorithm performance.

\begin{table}
\centering
\caption{Comparison between MBPO with and without model reset under different model UTD ratios. We show the point estimates and 95\% confidence interval. The results are evaluated over 5 random seeds.}
\begin{subtable}{\columnwidth}
\centering
\begin{tabular}{l c c c c}
\toprule
\diagbox{Reset}{Return}{Task} & HalfCheetah & Hopper & Walker2d & Humanoid \\
\midrule
without reset & \textbf{10893}(10077,11231) & 3455(3378,3588) & \textbf{3812}(2938,4324) & \textbf{4012}(3885,4311) \\
with reset & 10024(10003,11032) & \textbf{3588}(3471,3631) & 3612(2435,3921) & 3123(2683,3312) \\
\bottomrule
\end{tabular}

\caption*{(a) Model UTD ratio = 1$\times$dUTD}

\end{subtable}

\begin{subtable}{\columnwidth}
\centering
\begin{tabular}{l c c c c}
\toprule
\diagbox{Reset}{Return}{Task} & HalfCheetah & Hopper & Walker2d & Humanoid \\
\midrule
without reset & \textbf{11093}(10745,11543) & 3345(2878,3744) & \textbf{4133}(3974,4275) & \textbf{4452}(4382,4521) \\
with reset & 10937(10023,11832) & \textbf{3643}(3526,3825) & 3910(3879,3984) & 3904(3895,3921) \\
\bottomrule
\end{tabular}
\caption*{(b) Model UTD ratio = 2$\times$dUTD}
\end{subtable}

\begin{subtable}{\columnwidth}
\centering
\begin{tabular}{l c c c c}
\toprule
\diagbox{Reset}{Return}{Task} & HalfCheetah & Hopper & Walker2d & Humanoid \\
\midrule
without reset & \textbf{11783}(10121,11984) & 2374(1143,3792) & 3997(3849,4210) & 3546(2478,3901) \\
with reset & 11393(11359,12004) & \textbf{3781}(2988,4206) & \textbf{4128}(4011,4341) & \textbf{4321}(3894,4784) \\
\bottomrule
\end{tabular}
\caption*{(c) Model UTD ratio = 5$\times$dUTD}
\end{subtable}

\begin{subtable}{\columnwidth}
\centering
\begin{tabular}{l c c c c}
\toprule
\diagbox{Reset}{Return}{Task} & HalfCheetah & Hopper & Walker2d & Humanoid \\
\midrule
without reset & \textbf{11847}(10384,12483) & 2631(1935,3120) & 2398(943,3782) & 3135(2916,3289) \\
with reset & 11748(11543,11894) & \textbf{3123}(3024,3361) & \textbf{3266}(2411,4053) & \textbf{4104}(4023,4254) \\
\bottomrule
\end{tabular}
\caption*{(d) Model UTD ratio = 10$\times$dUTD}
\end{subtable}

    \label{tab:9}
\end{table}

\begin{table}
\centering
\caption{Comparison of final performance between DreamerV2 with and without model reset on DMC under different model UTD ratios. We compare IQM, Mean and Median score aggregated from 20 tasks on DMC. The results are evaluated by 5 runs.}
\begin{subtable}{\columnwidth}
\centering
\begin{tabular}{l c c c}
\toprule
\diagbox{Reset}{Score}{Metric} & IQM & Mean & Median \\
\midrule
without reset & \textbf{687}(669,720) & \textbf{702}(671,719) & \textbf{672}(652,698) \\
with reset & 633(621,649) & 644(630,652) & 639(627,660)\\
\bottomrule
\end{tabular}

\caption*{(a) Model UTD ratio = 1$\times$dUTD}

\end{subtable}

\begin{subtable}{\columnwidth}
\centering
\begin{tabular}{l c c c}
\toprule
\diagbox{Reset}{Score}{Metric} & IQM & Mean & Median \\
\midrule
without reset & 617(604,629) & 631(597,661) & 632(611,649) \\
with reset & \textbf{700}(674,728) & \textbf{709}(688,732) & \textbf{676}(670,691)\\
\bottomrule
\end{tabular}

\caption*{(b) Model UTD ratio = 5$\times$dUTD}

\end{subtable}

\begin{subtable}{\columnwidth}
\centering
\begin{tabular}{l c c c}
\toprule
\diagbox{Reset}{Score}{Metric} & IQM & Mean & Median \\
\midrule
without reset & 534(509,551) & 620(594,644) & 615(600,639) \\
with reset & \textbf{673}(660,687) & \textbf{678}(669,685) & \textbf{674}(666,680)\\
\bottomrule
\end{tabular}

\caption*{(c) Model UTD ratio = 10$\times$dUTD}

\end{subtable}

\label{tab:10}
\end{table}

\begin{figure}
\centering
\includegraphics[width=\columnwidth]{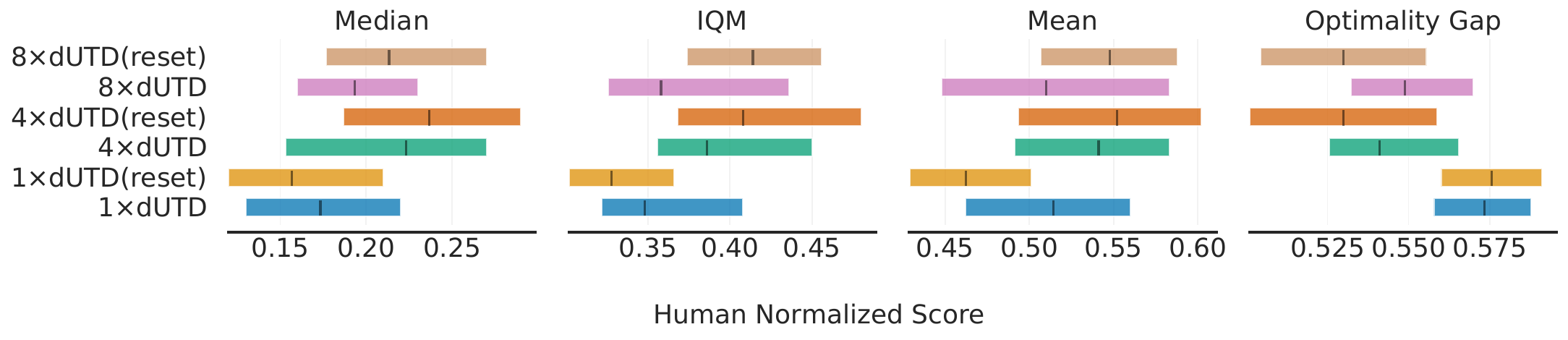}
\caption{Comparison between DreamerV2 with and without model reset under different model UTD ratios. We plot the point estimates and 95\% confidence interval of aggregated metrics on 26 tasks of Atari 100k. The results are evaluated by 5 runs.} \label{fig:15}
\end{figure}

\subsection{The impact of model ensemble size}
\begin{figure}[htb]
\centering
\includegraphics[width=\columnwidth]{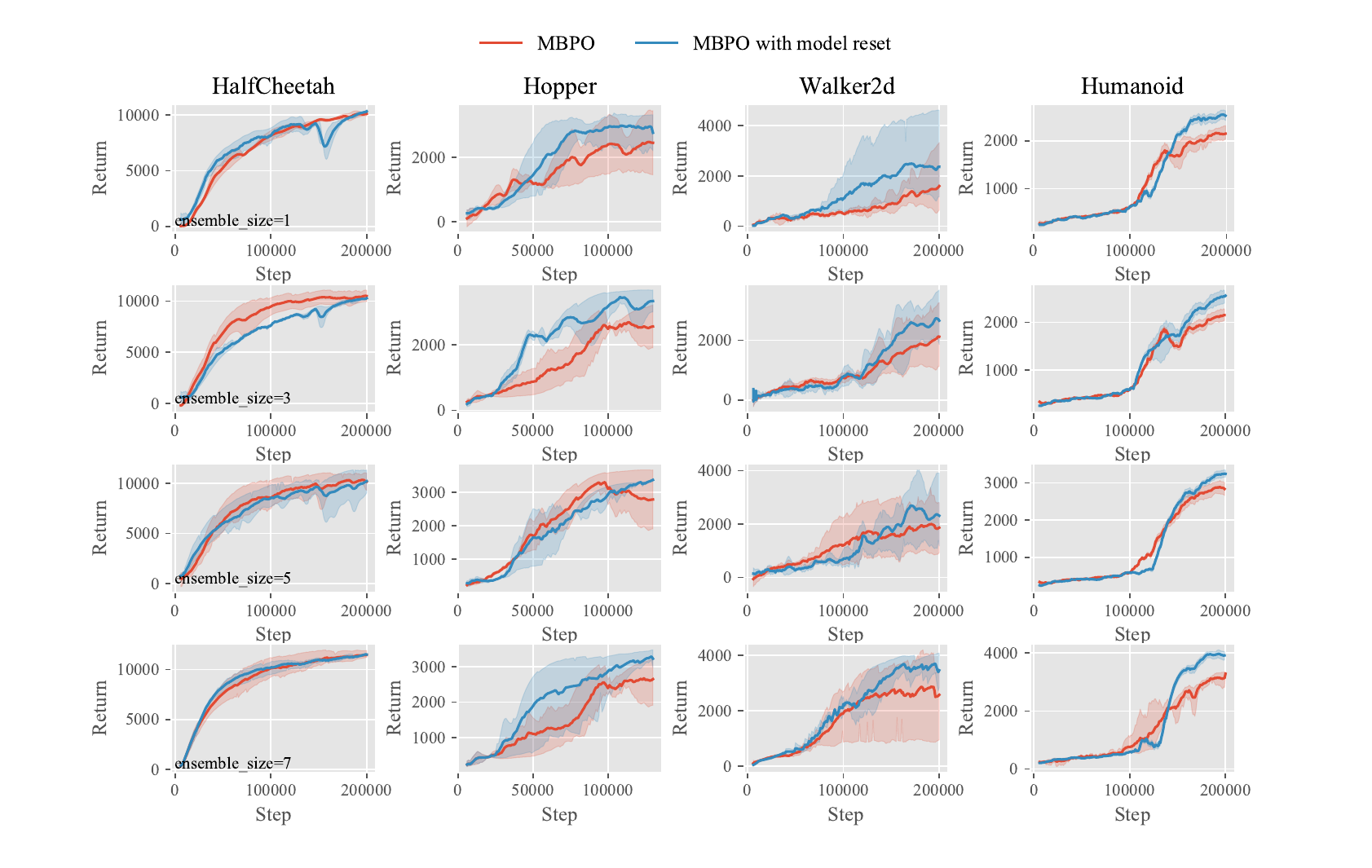}
\caption{Learning curves for MBPO and MBPO with different ensemble sizes. The results are evaluated by 5 runs.} \label{fig:10}
\end{figure}

This subsection provides more experimental results on how model ensemble size affects the primacy bias and the effect of model reset. We conduct experiments on MBPO and MBPO with model reset. We conduct experiments on four tasks of MuJoCo, where model UTD ratio is set to 10, and model ensemble size is varied as 1, 3, 5, and 7. Figure \ref{fig:10} shows the learning curves of MBPO with and without model reset with different model ensemble sizes. Table \ref{tab:11} compares final performance between MBPO with and without model reset with different model ensemble sizes. Table \ref{tab:11} also demonstrates the extent to which model reset improves the final performance on different tasks with varying model ensemble sizes. Apart from HalfCheetah-v2 task, model reset has almost no impact on performance. In the overall trend, we can see that as model ensemble size increases, the performance improvement brought about by model reset tends to decrease for the other three tasks. However, it is also evident that the variation is not consistently monotonic. For Hopper-v2 task, the performance improvement from model reset increases significantly when the model ensemble size is 3. For Walker2d-v2 and Humanoid-v2 tasks, the performance improvement from model reset increases when the model ensemble size is 7. We assume this is because a larger model ensemble size raises the performance limit of the algorithm, allowing model reset to have a more significant impact. We leave further exploration of this phenomenon for future work.

\begin{table}
\centering
\caption{Performance comparison between MBPO and MBPO with model reset, considering different ensemble sizes. We also include the performance improvement achieved by using model reset.}
\begin{subtable}{\columnwidth}
\centering
\begin{tabular}{l c c c c}
\toprule
\diagbox{Reset}{Return}{Task} & HalfCheetah & Hopper & Walker2d & Humanoid \\
\midrule
without reset & 9878(9805,9927) & 2374(1644,3389) & 1722(874,3489) & 2034(1975,2189) \\
with reset & \textbf{10018}(9980,10014) & \textbf{2516}(2138,2989) & \textbf{2512}(1435,4421) & \textbf{3123}(2683,3312) \\
\midrule
Improvement & 1.4\% & 6.0\% & 45.9\% & 53.5\%\\
\bottomrule
\end{tabular}

\caption*{(a) ensemble size = 1}

\end{subtable}

\begin{subtable}{\columnwidth}
\centering
\begin{tabular}{l c c c c}
\toprule
\diagbox{Reset}{Return}{Task} & HalfCheetah & Hopper & Walker2d & Humanoid \\
\midrule
without reset & \textbf{10093}(9645,11043) & 2441(1907,3157) & 2133(1074,2901) & 2253(2041,2113) \\
with reset & 10024(9983,10384) & \textbf{3623}(3378,3896) & \textbf{2841}(1879,3788) & \textbf{2527}(2389,2760) \\
\midrule
Improvement & -0.68\% & 48.4\% & 33.2\% & 12.2\%\\
\bottomrule
\end{tabular}
\caption*{(b) ensemble size = 3}
\end{subtable}

\begin{subtable}{\columnwidth}
\centering
\begin{tabular}{l c c c c}
\toprule
\diagbox{Reset}{Return}{Task} & HalfCheetah & Hopper & Walker2d & Humanoid \\
\midrule
without reset & \textbf{10183}(8979,11244) & 2811(1908,3801) & 1977(1159,2958) & 2898(2801,3040) \\
with reset & 10014(9043,12013) & \textbf{3342}(3207,3398) & \textbf{2301}(1309,3928) & \textbf{3218}(3178,3301) \\
\midrule
Improvement & -1.7\% & 18.9\% & 16.4\% & 11.0\%\\
\bottomrule
\end{tabular}
\caption*{(c) ensemble size = 5}
\end{subtable}

\begin{subtable}{\columnwidth}
\centering
\begin{tabular}{l c c c c}
\toprule
\diagbox{Reset}{Return}{Task} & HalfCheetah & Hopper & Walker2d & Humanoid \\
\midrule
without reset & \textbf{11847}(10384,12483) & 2631(1935,3120) & 2398(943,3782) & 3135(2916,3289) \\
with reset & 11748(11543,11894) & \textbf{3123}(3024,3361) & \textbf{3266}(2411,4053) & \textbf{4104}(4023,4254) \\
\midrule
Improvement & -0.84\% & 18.7\% & 36.2\% & 30.9\%\\
\bottomrule
\end{tabular}
\caption*{(d) ensemble size = 7}
\end{subtable}

    \label{tab:11}
\end{table}

\subsection{When and how to apply model reset} 
In this subsection, we present experimental results to support our opinion in the main text regarding when and how to reset the model. Our experiments are based on MBPO. 

\begin{figure}[htb] 
\centering
\includegraphics[width=\columnwidth]{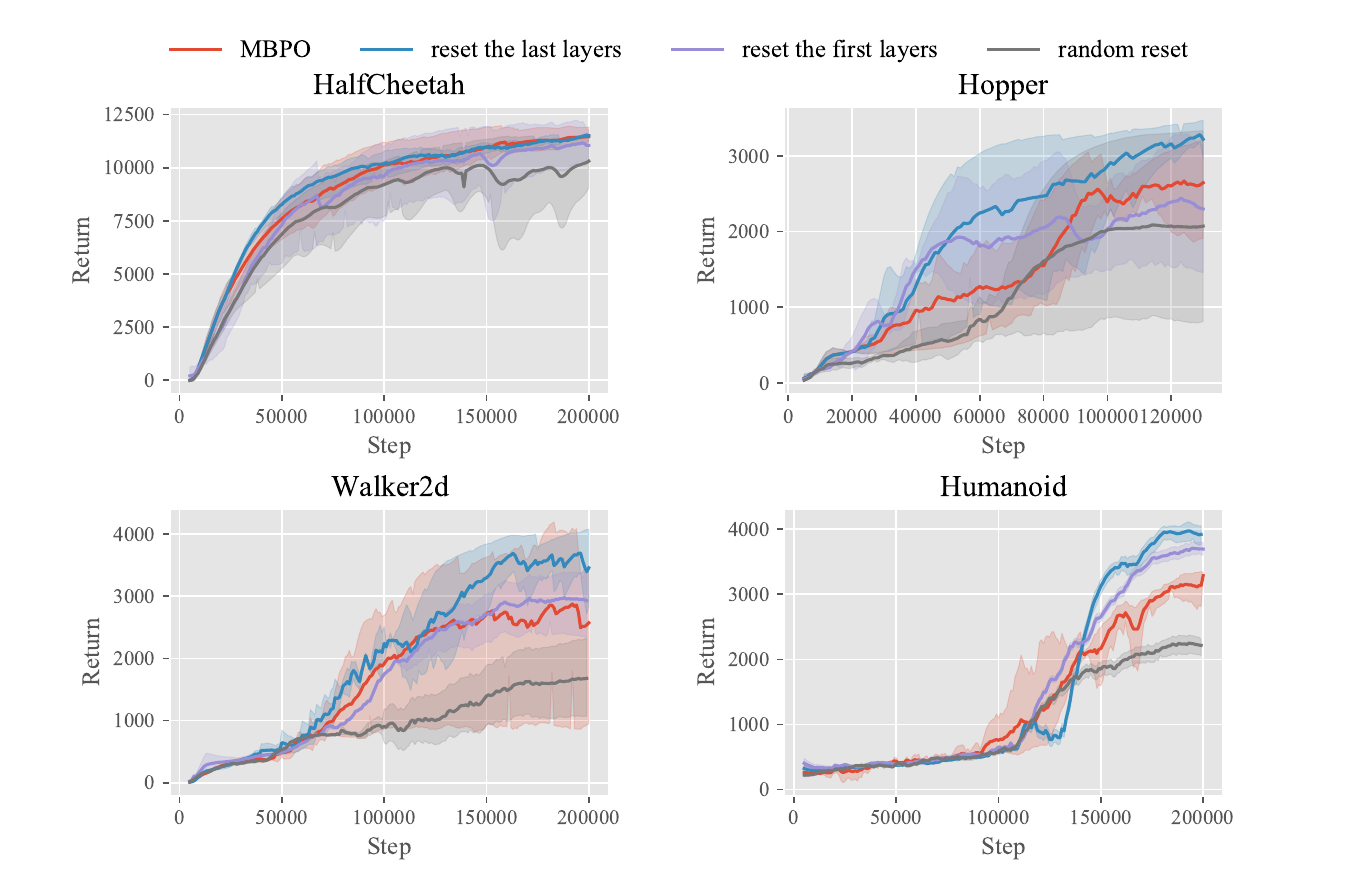}
\caption{Learning curves of MBPO and MBPO with different parts of the network reset. The results are evaluated by 5 runs.} \label{fig:11}
\end{figure}

First, we answer the question of which part of the network should be reset. We set model UTD ratio to 10 and try three model reset strategies: resetting only the first two layers of the network, resetting only the last two layers, and randomly resetting two layers. Figure \ref{fig:11} and Table \ref{tab:12} show the learning curves and results of these three reset strategies. As shown, resetting only the last two layers of the network achieves the highest performance improvement, while resetting the first two layers of the network and random resetting may even potentially harm the performance of the original algorithm.

\begin{figure}[htb] 
\centering
\includegraphics[width=\columnwidth]{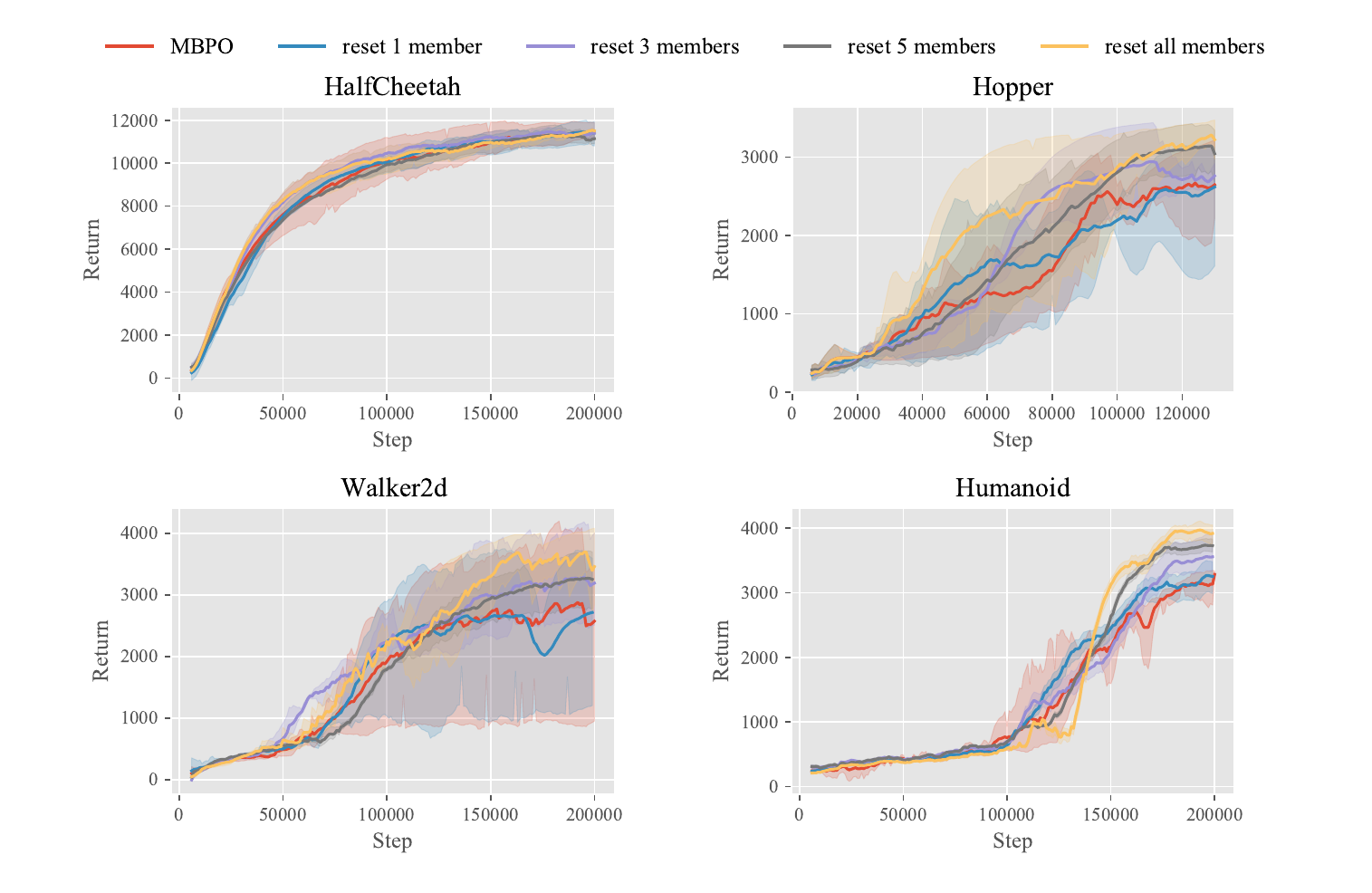}
\caption{Learning curves of MBPO and MBPO with different numbers of reset ensemble members. The results are evaluated by 5 runs.} \label{fig:12}
\end{figure}

Second, we want to know if resetting only a portion of the ensemble members has a better or worse effect on the performance than resetting all the ensemble members. To answer this question, we set model UTD ratio to 10 and reset a different number of ensemble members, specifically 1, 3, 5, and all 7 members individually. We record the learning curves and results in Figure \ref{fig:12} and Table \ref{tab:13}. The experimental results indicate that apart from the HalfCheetah task, not performing any resets yields the best performance. However, for the other three tasks, resetting all the ensemble members yields the best performance. This is easily understandable as all the ensemble members may suffer from primacy bias. Therefore, resetting all the ensemble members is the most effective strategy.

\begin{figure}[htb] 
\centering
\includegraphics[width=\columnwidth]{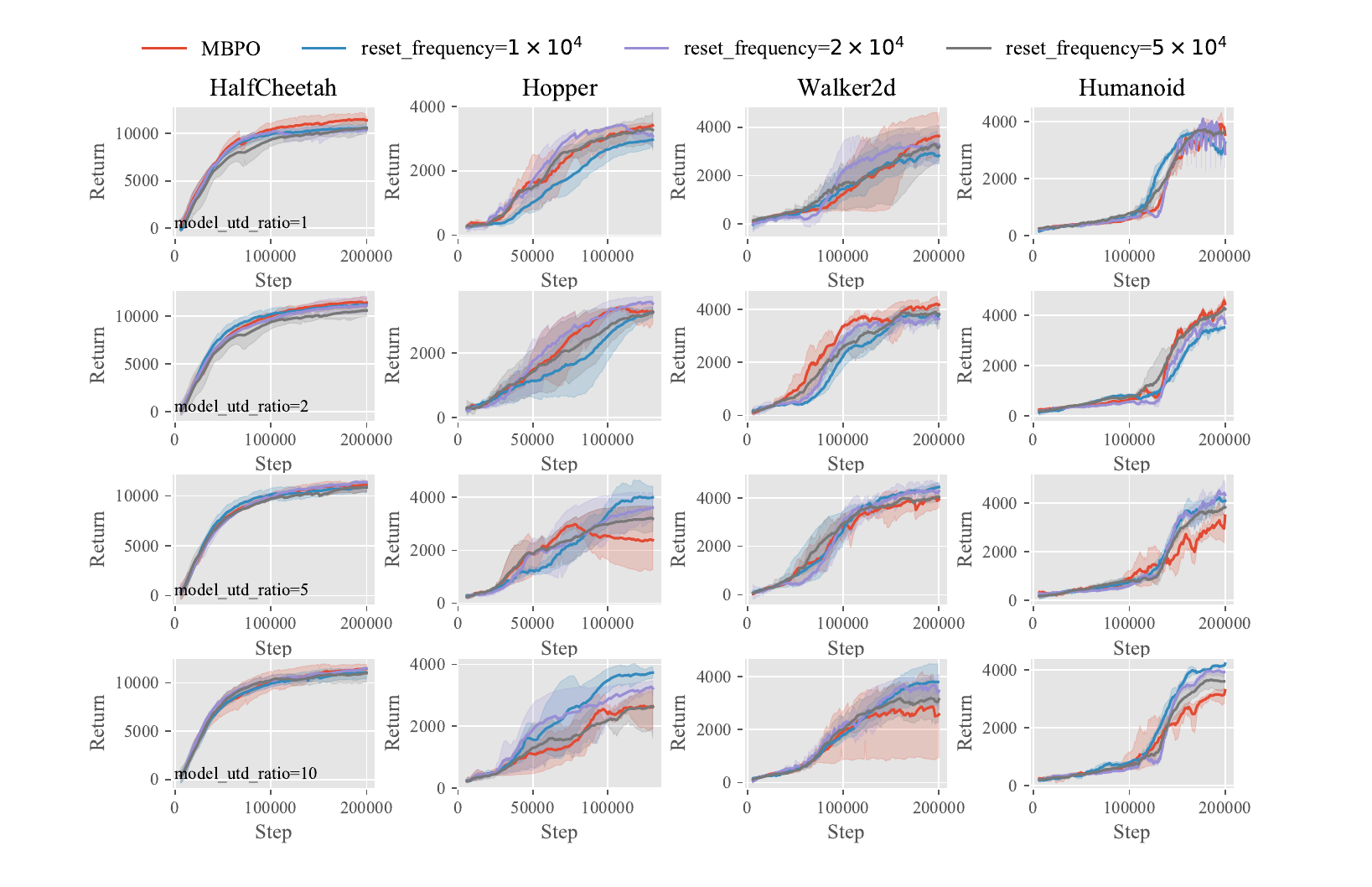}
\caption{Learning curves for MBPO and MBPO with different reset intervals on four tasks of MuJoCo. The results are evalated by 5 runs.} \label{fig:13}
\end{figure}

Last, we want to investigate the appropriate frequency for performing model reset. Since the effectiveness of model reset is largely influenced by model UTD ratio, we naturally assume that as model UTD ratio increases, the impact of the primacy bias becomes more pronounced. Thus the frequency of resets should be higher. In our experiment, we set different reset frequency (perform a reset every $1\times10^4$steps, $2\times10^4$ steps and $5\times10^4$ steps) for different model UTD ratios (1, 5, and 10). We present the learning curves and experimental results in Figure \ref{fig:13} and Table \ref{tab:14}. The experimental results generally support our assumption. As model UTD ratio increases, moderately increasing the frequency of resets leads to better performance.

\begin{table}
\centering
\caption{Performance comparison between MBPO and MBPO with different parts of the network reset. The results are evaluated by 5 runs.}
\begin{tabular}{l c c c c}
\toprule
\diagbox{Reset}{Return}{Task} & HalfCheetah & Hopper & Walker2d & Humanoid \\
\midrule
without reset & \textbf{11847}(10384,12483) & 2631(1935,3120) & 2398(943,3782) & 3135(2916,3289) \\
reset first layers & 11523(11072,12079) & 2241(1519,3146) & 2978(2386,3479) & 3749(3617,3879) \\
reset last layers & 11748(11543,11894) & \textbf{3123}(3024,3361) & \textbf{3266}(2411,4053) & \textbf{4104}(4023,4254) \\
random reset & 10274(8476,11793) & 2183(894,3271) & 1748(1103,2375) & 2145(2017,2318) \\
\bottomrule
\end{tabular}

\label{tab:12}
\end{table}

\begin{table}
\centering
\caption{Performance comparison between MBPO and MBPO with different numbers of reset ensemble members. The results are evaluated by 5 runs.}
\begin{tabular}{l c c c c}
\toprule
\diagbox{Reset}{Return}{Task} & HalfCheetah & Hopper & Walker2d & Humanoid \\
\midrule
without reset & \textbf{11847}(10384,12483) & 2631(1935,3120) & 2398(943,3782) & 3135(2916,3289) \\
reset 1 member & 11523(11072,12079) & 2647(1688,3371) & 2817(1249,3718) & 3301(3006,3538) \\
reset 3 members & 11525(10938,11974) & 2718(2481,3364) & 3219(1283,3504) & 3571(3210,3741) \\
reset 5 members & 11501(11473,11892) & 3018(2741,3201) & 3256(2684,3658) & 3786(3689,3923) \\
reset 7 members & 11748(11543,11894) & \textbf{3123}(3024,3361) & \textbf{3266}(2411,4053) & \textbf{4104}(4023,4254) \\
\bottomrule
\end{tabular}

\label{tab:13}
\end{table}

\begin{table}
\centering
\caption{Performance comparison of MBPO and MBPO with different reset intervals. The results are evaluated by 5 runs.}

\begin{subtable}{\columnwidth}
\centering
\begin{tabular}{l c c c c}
\toprule
\diagbox{Reset}{Return}{Task} & HalfCheetah & Hopper & Walker2d & Humanoid \\
\midrule
without reset & \textbf{10893}(10077,11231) & 3455(3378,3588) & \textbf{3812}(2938,4324) & \textbf{4012}(3885,4311) \\
$1\times10^4$ & 10031(9917,11231) & 3123(2980,3310) & 2648(2389,3472) & 2736(2699,2868) \\
$2\times10^4$ & 10024(10003,11032) & \textbf{3588}(3471,3631) & 3612(2435,3921) & 3123(2683,3312) \\
$5\times10^4$ & 9974(9910,10214) & 3401(2847,3621) & 3470(2231,3980) & 3804(3681,3931) \\
\bottomrule
\end{tabular}

\caption*{(a) Model UTD ratio = 1$\times$dUTD}

\end{subtable}

\begin{subtable}{\columnwidth}
\centering
\begin{tabular}{l c c c c}
\toprule
\diagbox{Reset}{Return}{Task} & HalfCheetah & Hopper & Walker2d & Humanoid \\
\midrule
without reset & \textbf{11093}(10745,11543) & 3345(2878,3744) & \textbf{4133}(3974,4275) & \textbf{4452}(4382,4521) \\
$1\times10^4$ & 10897(10024,11379) & 3288(3074,3348) & 3961(3811,4132) & 3617(3578,3744) \\
$2\times10^4$ & 10937(10023,11832) & \textbf{3643}(3526,3825) & 3910(3879,3984) & 3904(3895,3921) \\
$5\times10^4$ & 10217(10039,12741) & 3319(3104,3581) & 3712(3411,4012) & 4312(4156,4482) \\
\bottomrule
\end{tabular}
\caption*{(b) Model UTD ratio = 2$\times$dUTD}
\end{subtable}

\begin{subtable}{\columnwidth}
\centering
\begin{tabular}{l c c c c}

\toprule
\diagbox{Reset}{Return}{Task} & HalfCheetah & Hopper & Walker2d & Humanoid \\
\midrule
without reset & \textbf{11783}(10121,11984) & 2374(1143,3792) & 3997(3894,4210) & 3546(2478,3901) \\
$1\times10^4$ & 10103(9701,10421) & \textbf{3970}(3771,4310) & \textbf{4225}(3971,4369) & 4032(3897,4183) \\
$2\times10^4$ & 11393(11359,12004) & 3781(2988,4206) & 4128(4011,4341) & \textbf{4321}(3894,4784) \\
$5\times10^4$ & 10174(10071,11014) & 3148(2719,3782) & 4051(3908,4182) & 3878(3791,3948) \\
\bottomrule
\end{tabular}
\caption*{(c) Model UTD ratio = 5$\times$dUTD}
\end{subtable}

\begin{subtable}{\columnwidth}
\centering
\begin{tabular}{l c c c c}

\toprule
\diagbox{Reset}{Return}{Task} & HalfCheetah & Hopper & Walker2d & Humanoid \\
\midrule
without reset & \textbf{11847}(10384,12483) & 2631(1935,3120) & 2398(943,3782) & 3135(2916,3289) \\
$1\times10^4$ & 11385(10472,12014) & \textbf{3807}(3711,3965) & \textbf{3941}(3298,4487) & \textbf{4261}(4197,4271) \\
$2\times10^4$ & 11748(11543,11894) & 3123(3024,3361) & 3266(2411,4053) & 4104(4023,4254) \\
$5\times10^4$ & 11385(10459,11814) & 2651(1910,3208) & 2985(2471,3721) & 3618(3172,4072) \\
\bottomrule
\end{tabular}
\caption*{(d) Model UTD ratio = 10$\times$dUTD}
\end{subtable}

    \label{tab:14}
\end{table}

\end{document}